\documentclass[10pt,twocolumn]{article}
\usepackage{simpleConference}

\usepackage{times}

\usepackage{graphicx}
\usepackage{amsmath}
\usepackage{amssymb}
\usepackage{booktabs}

\usepackage{bm}
\usepackage{enumerate}
\usepackage{multirow}
\usepackage{array}
\usepackage{arydshln}
\usepackage{makecell}

\usepackage[algo2e]{algorithm2e}
\usepackage{amsfonts}       % blackboard math symbols
\usepackage{nicefrac}       % compact symbols for 1/2, etc.
\usepackage{microtype}      % microtypography

\usepackage{animate}
\usepackage{xcolor}
\usepackage{framed}   
\usepackage{algorithm}
\usepackage{algorithmic}
\newcolumntype{C}[1]{>{\centering\let\newline\\\arraybackslash}m{#1}}
\usepackage{appendix}
\usepackage{lipsum}
\usepackage{caption}
\usepackage[pagebackref,breaklinks,colorlinks]{hyperref}

\usepackage[capitalize]{cleveref}
\crefname{section}{Sec.}{Secs.}
\Crefname{section}{Section}{Sections}
\Crefname{table}{Table}{Tables}
\crefname{table}{Tab.}{Tabs.}

\begin{document}

\title{HS-Diffusion: Semantic-Mixing Diffusion for Head Swapping}

\author{Qinghe~Wang$^{1\dagger}$, Lijie~Liu$^{2}$, Miao~Hua$^{2}$, Pengfei~Zhu$^{3}$, Wangmeng Zuo$^{4}$, \\Qinghua~Hu$^{3}$, Huchuan~Lu$^{1}$, Bing~Cao$^{3}$\\
$^1$Dalian University of Technology, $^2$ByteDance Inc, $^3$Tianjin University, $^4$Harbin Institute of Technology
}

\twocolumn[{
\renewcommand\twocolumn[1][]{#1}
\maketitle  
\vspace{-0.65cm}
\begin{center}
\centering
    \captionsetup{type=figure}
    \includegraphics[width=1\linewidth]{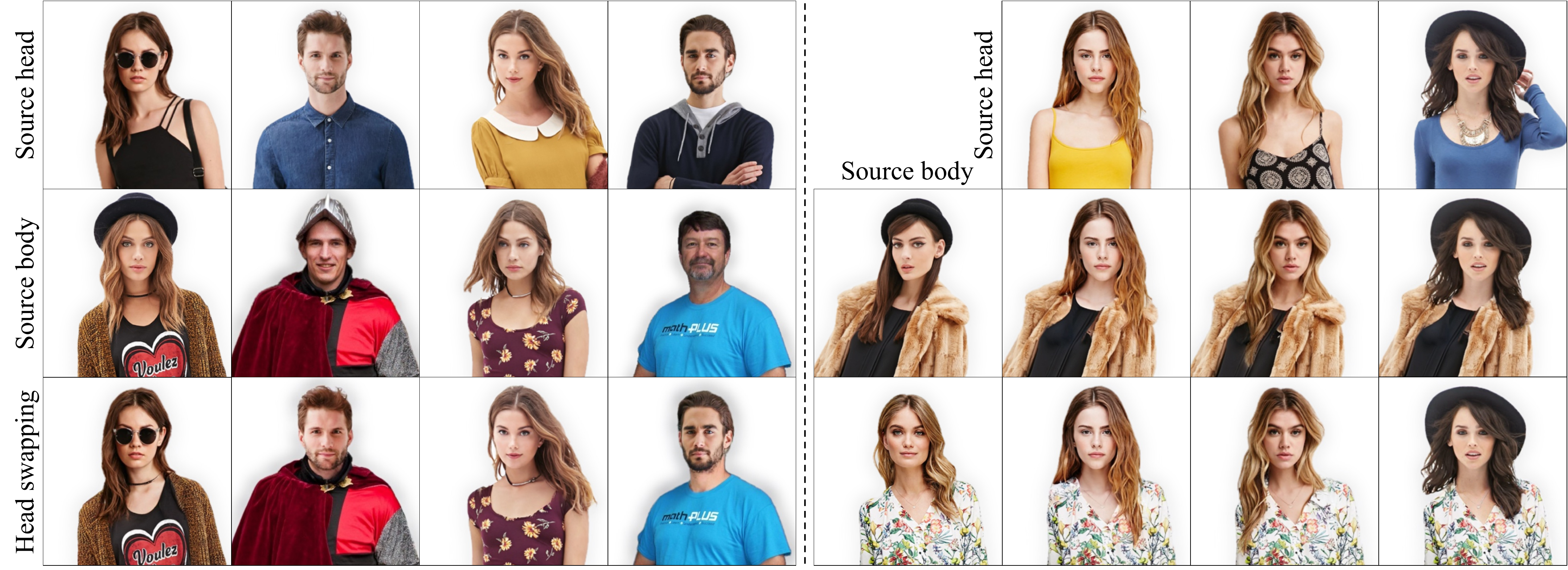}
    \captionof{figure}{Head swapping results generated by our framework. The left shows that the source heads and source bodys are preserved flawlessly and the transition regions are inpainted seamlessly. The right further demonstrates the effectiveness of our framework, where a source head can be paired with various bodys and produce high-quality head swapping results, and vice versa.}
    \label{fig:first_image}
\end{center}
}]

\newcommand\blfootnote[1]{%
\begingroup
\renewcommand\thefootnote{}\footnote{#1}%
\addtocounter{footnote}{-1}%
\endgroup
}

\blfootnote{$\dagger$Work done during an internship at ByteDance.}

%%%%%%%%% ABSTRACT
\begin{abstract}
\vspace{-0.37cm}
Image-based head swapping task aims to stitch a source head to another source body flawlessly. This seldom-studied task faces two major challenges: 1) Preserving the head and body from various sources while generating a seamless transition region. 2) No paired head swapping dataset and benchmark so far. In this paper, we propose a semantic-mixing diffusion model for head swapping (HS-Diffusion) which consists of a latent diffusion model (LDM) and a semantic layout generator. We blend the semantic layouts of source head and source body, and then inpaint the transition region by the semantic layout generator, achieving a coarse-grained head swapping. Semantic-mixing LDM can further implement a fine-grained head swapping with the inpainted layout as condition by a progressive fusion process, while preserving head and body with high-quality reconstruction. To this end, we propose a semantic calibration strategy for natural inpainting and a neck alignment for geometric realism. Importantly, we construct a new image-based head swapping benchmark and design two tailor-designed metrics (Mask-FID and Focal-FID). Extensive experiments demonstrate the superiority of our framework. The code will be available: \textcolor{magenta}{\href{https://github.com/qinghew/HS-Diffusion}{https://github.com/qinghew/HS-Diffusion}}.
\end{abstract}

\section{Introduction}
\label{sec:intro}
Recent advances in generative adversarial networks~\cite{fu2022stylegan,zhao2020layout2image} and diffusion models~\cite{rombach2022high,saharia2022photorealistic,dhariwal2021diffusion} have exhibited unprecedented success in generative and editing applications~\cite{tov2021designing,zhang2023adding}. Numerous emerged methods that edit with two images focus on style transfer~\cite{lee2020drit++}, face swapping~\cite{chen2020simswap}, and attribute transfer~\cite{georgopoulos2021mitigating} by information transmission, but few have explored the physiological blending of two images. In this paper, we propose an image-based head swapping task that aims to perform a large-scale replacement of the entire head onto the target body, while maintaining the main components of the two source images unchanged. As shown in Fig.~\ref{fig:first_image}, our framework seamlessly stitches the source head with another source body. Not only are heads and bodies flawlessly preserved, but also \textbf{the transition region}~(including the neck region and the covered region by the hair from source body image) is seamlessly inpainted. The success of head swapping task has important implications for a variety of applications in commercial and entertainment scenarios, such as virtual try-on~\cite{han2018viton,li2021toward,yang2020towards} and portrait generation~\cite{liu2022semantic, su2022drawinginstyles,Li_Huang_Cao_He_Tan_2020}.

Although great progress has been made in face swapping task~\cite{perov2020deepfacelab,xu2022high,xu2022region}, head swapping task has not been widely studied yet, especially the image-based head swapping. Face swapping only needs to transfer the identity information of source image to another image, but head swapping requires migrating a much larger region (i.e., face and hair) and considering the gap between various people. And the transition region needs to be seamlessly inpainted without artifacts. However, there is no paired head swapping dataset and no method designed for the image-based head swapping so far. Besides, the existing alignment technique~\cite{kazemi2014one, fu2022stylegan} can align the face or body, but cannot solve the horizontal deviations for the heads with different face orientations, which causes more difficulties. An auxiliary pre-processing operation is to employ the SOTA image animation~\cite{wang2022latent, zhang2023sadtalker} or 3D GAN~\cite{chan2022efficient, zhang2023multi, Moschoglou_Ploumpis_Nicolaou_Papaioannou_Zafeiriou_2020} techniques to align head pose and before head swapping, but it still needs to deal with the unavailable transition to connect the head and body from different sources. Moreover, it complicates the pipeline and might result in detail compromise and artifacts. Therefore, without changing the pose, we focus to implement head swapping with two source images straightforwardly, which faces huge challenges.

To tackle this issue, we propose a semantic-mixing diffusion model (HS-Diffusion) for coarse-to-fine head swapping which consists of a latent diffusion model (LDM) and a semantic layout generator. For better control of the background region and the transition region with diverse classes, we choose the semantic layout as condition to train LDM. However, it is difficult to directly obtain available semantic layouts for head swapping, so we design a semantic layout generator to implement a coarse-grained head swapping at semantic level. Inspired by the text-driven diffusion~\cite{avrahami2022blended_0,avrahami2022blended_1}, we propose to mix the diffusion latents of the transition region with the noises of source head and source body at each noising level under semantic guidance. This progressive fusion process iteratively fills the inpainted semantic layout with suitable textures and harmonizes the transition region with surroundings, which implements a fine-grained head swapping at pixel level. To this end, we propose a semantic calibration strategy by training with a head-cover augmentation which enables the semantic layout generator to inpaint and calibrate the transition region of layout. It also allows LDM to calibrate the details and desensitize to semantic noise. In addition, we propose a neck alignment trick to solve the problem that face-aligned images may lead to unrealistic head swapping results due to the neck misalignment. Furthermore, we construct a new image-based head swapping benchmark and propose two tailor-designed evaluation metrics (Mask-FID and Focal-FID). We also implement several baselines for head swapping and compare them with our proposed HS-Diffusion on this benchmark. Extensive experiments show that our framework is effective and prominent. We hope this benchmark will help the community and advance image-based head swapping research.

In summary, our contributions are three-fold:
\begin{itemize}
    \item We first propose a semantic-mixing diffusion model for head swapping, which blends the semantic layouts to guide the mixing of diffusion latents step-by-step, stitching one head to another body seamlessly.
    
    \item We propose a semantic calibration strategy to adaptively inpaint incomplete region and address the occlusion and noise issues encountered for head swapping.
    
    \item We develop a plug-and-play neck alignment to improve geometric realism for downstream models and two variants of FID for evaluation. Extensive experiments demonstrate the superiority of our framework. As a new image-based head swapping benchmark, our code will be publicly available.
\end{itemize}

The remaining paper is organized as follows. In Sec.~\ref{related_work}, we briefly summarize the related work. In Sec.~\ref{sec:method}, we introduce the proposed head swapping framework in detail. Extensive experiments are conducted in Sec.~\ref{sec:experiment} to evaluate the performance of our framework, with in-depth analysis. Sec.~\ref{sec:discussion} further discusses more applications and effects of modules. Sec.~\ref{sec:conclusion} concludes this work.

\section{Related work}
\label{related_work}

\subsection{Face and Head Swapping}
\label{related_work_editing}
Face swapping~\cite{zhu2021one, xu2021facecontroller, kim2022smooth, xu2022high, xu2022region} is a popular task and has been widely applied for digital entertainment in recent years. These methods transfer the identity representation from source image to target image~\cite{li2019faceshifter, chen2020simswap}, without concern for the other characteristics of source image, such as face shape and hairstyle. In contrast, the head swapping task is more difficult, but has rarely been studied so far. StylePoseGAN~\cite{albahar2021pose} is designed for re-rendering with pose/appearance and presents a few head swapping samples, but suffers from identity ambiguity. HeSer~\cite{shu2022few} is the first work to implement few-shot head swapping, however, it needs videos data to migrate the head pose and ignores the mentioned transition region issue. In this paper, we propose a new image-based head swapping framework to fill the gaps in previous research.

A naive way to perform the image-based head swapping is to cut the source head and another source body$+$neck, and then paste them on a canvas, but the incomplete transition region makes the result unrealistic. Recently developed deep generative models~\cite{jabbar2021survey,xia2022gan} have the potential to solve this problem. The GAN-based inpainting methods~\cite{isogawa2019better,liu2021pd,li2022mat} might be able to inpaint unavailable regions and fusion with surroundings. PDGAN~\cite{liu2021pd} incorporates context constraint by modulating deep random noise features with SPDNorm for inpainting. MAT~\cite{li2022mat} designs a multi-head contextual attention to exploit valid tokens with a dynamic mask to inpaint missing region stably. Besides, the latent-space editing methods~\cite{kim2021exploiting, fruhstuck2022insetgan} also have the potential to achieve the head swapping by fusing the latent codes.
The encoder-based method StyleMapGAN~\cite{kim2021exploiting} can conduct the semantic-guided manipulation with the spatial latent codes for face images, which might work on our half-body dataset as well. The optimization-based method InsetGAN~\cite{fruhstuck2022insetgan} can implement head swapping between the generated images by optimizing the latent codes with a face StyleGAN2~\cite{karras2020analyzing} and a half-body StyleGAN2. Assisted with the inversion methods~\cite{karras2019style,tov2021designing}, InsetGAN also has the potential capability of head swapping for real images. 
To summarize, the image-level inpainting methods~\cite{liu2021pd,li2022mat} have near-perfect reconstruction ability but suffer from unrealistic stitching results. The latent-space fusion methods~\cite{kim2021exploiting,fruhstuck2022insetgan} achieve natural fusion but have poor reconstruction performance. Fortunately, our diffusion-based method can seamlessly fusion the source head and source body with the generated transition region while preserving with high-quality reconstruction.

\subsection{Denoising Diffusion Probabilistic Models}
\label{related_work_DDPM}
Recently, Denoising Diffusion Probabilistic Models (DDPMs)~\cite{ho2020denoising, nichol2021improved, dhariwal2021diffusion, graikos2022diffusion} have achieved amazing performance for image generation and attracted increasing attention. DDPMs can progressively add Gaussian noise to an input image $x_0$ to $x_t$ with variance $\beta_t\in(0,1)$ at time $t\in\{0,1,\cdots T\}$ by $q(x_t\vert x_{t-1}) = \mathcal{N}(\sqrt{1-\beta_t}x_{t-1},\beta_t\mathbf{I})$. This forward noising process further can directly sampled from $x_0$ without the intermediate steps:
\begin{equation}\label{forward_noise}
\begin{aligned}
q(x_t\vert x_0) = \mathcal{N}(\sqrt{\bar{\alpha}_{t}}x_0,(1-\bar{\alpha}_t)\mathbf{I})\\
x_t = \sqrt{\bar{\alpha}_{t}}x_0+\sqrt{1-\bar{\alpha}_t}\epsilon,
\end{aligned}
\end{equation}
where $\alpha_t=1-\beta_t,\bar{\alpha}_t=\prod_{s=0}^{t}\alpha_s$ and $\epsilon$ is randomly sampled from $\mathcal{N}(0,\mathbf{I})$. The reverse diffusion process $p_\theta(x_{t-1}\vert x_t)$ can be modeled as $\mathcal{N}(\mu_\theta(x_t,t),\sigma_t)$ with a neural network $\epsilon_\theta$ for predicting noise:
\begin{equation}
\label{L_dm}
     \mathcal{L}_{DM} = \mathbb{E}_{x,\epsilon\sim\mathcal{N}(0,\mathbf{I}),t}\left[\|\epsilon-\epsilon_\theta(x_t,t)\|^2_2\right]
\end{equation}
A random noise $x_T\in\mathcal{N}(0,\mathbf{I})$ can be denoised to an image by iterating the reverse diffusion process. Without changing the forward noising process, Denoising Diffusion Implicit Model (DDIM)~\cite{song2020denoising} further proposes to accelerate sampling:
\begin{equation}
\begin{small}  %  footnotesize
\begin{aligned}
\label{ddim_denoise}
x_{t-1} = &\sqrt{\alpha_{t-1}}\left(\frac{x_t-\sqrt{1-\alpha_t}\epsilon_\theta^{(t)}(x_t)}{\sqrt{\alpha_t}}\right)\\ &+ \sqrt{1-\alpha_{t-1} - \sigma_t^2}\cdot\epsilon_\theta^{(t)}(x_t) + \sigma_t\epsilon_t
\end{aligned}
\end{small} 
\end{equation}

DDIM shares both the objective and training process with DDPMs, and only runs faster for inference sampling. But they work in pixel space, which causes a huge computational cost~\cite{lu2022dpm}. Latent Diffusion Model~(LDM)~\cite{rombach2022high} demonstrates that diffusion models perform better in a low-dimensional latent space, as they bypass redundant information in pixel space and concentrate on the low-dimensional representation in latent space.

Furthermore, conditional DDPMs~\cite{dhariwal2021diffusion,avrahami2022blended_0,avrahami2022blended_1,wang2022semantic, zhang2023adding, goel2023pair} aim to control the generated images as desired with condition guidance. ADM~\cite{dhariwal2021diffusion} proposes to use gradients from a classifier as sampling's guidance for class-specific generation which beats the SOTA GAN-based method BigGAN~\cite{brock2018large} on FID~\cite{lucic2017gans} for the first time. The text-to-image diffusion models~\cite{rombach2022high, saharia2022photorealistic} can generate high-quality and imaginative images with a text prompt which have demonstrated unprecedented capabilities, but lack accurate controllability~\cite{zhang2023adding, li2023gligen, mou2023t2i} such as structure guidance. Based on the pretrained text-to-image model such as Stable Diffusion~(SD), ControlNet~\cite{zhang2023adding} clones the weights of SD to learn the additional conditions, such as keypoints, edge maps, etc., to enrich controllability. PAIR-Diffusion~\cite{goel2023pair} explicitly extracts the structure and appearance information to train the diffusion model in a conditional manner, which can independently edit the structure and appearance of each object. But for the local editing, these methods requires manual modification on the condition and struggle to maintain robustness with low-quality conditions~\cite{li2023gligen}. The success of these diffusion models inspires our work to implement head swapping with the proposed semantic-mixing LDM.

\begin{figure*}[!t]
  \centering
  \includegraphics[width=1\linewidth]{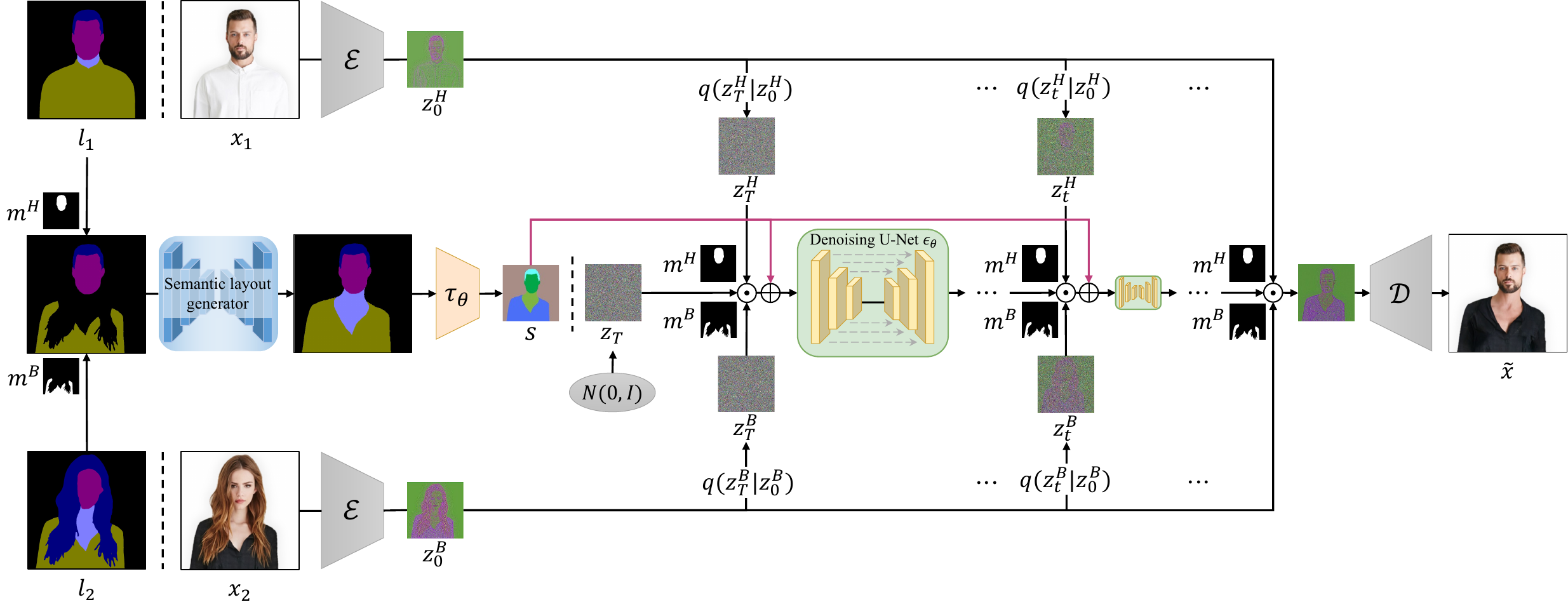}
      \caption{The image-based head swapping pipeline with our HS-Diffusion. We blend the semantic layout ($l_1,l_2$) with the head mask $m^H$ and body mask $m^B$, and then use the well-trained semantic layout generator to inpaint the incomplete transition region. A random noise $z_T$ sampled from $\mathcal{N}(0,\mathbf{I})$ will mix with $z_T^H$ and $z_T^B$ which are sampled from the forward noising process. The mixed noise will be concatenated with the semantic latent representation $s$ as the input of denoising U-Net $\epsilon_\theta$. We conduct the mixing and concatenation operations at each denoising step. Finally, we decode the $z_0$ to obtain a seamless head swapping result.}
  \label{fig:infer_process}
\end{figure*}

\section{Method}
\label{sec:method}
Given two half-body images~($x_1,x_2$) and the corresponding semantic layouts~($l_1,l_2$), we aim to produce a new fusion half-body image $\tilde{x}$ which preserves the head of $x_1$ and the body of $x_2$. Furthermore, the transition region should appear more seamless. To this end, we train a latent diffusion model (LDM) and a semantic layout generator separately which work together for head swapping. We summarize the image-based head swapping pipeline shown in Fig.~\ref{fig:infer_process} with the following steps: (i)~Blend the semantic layout ($l_1,l_2$) with the head mask $m^H$ and body mask $m^B$. (ii)~Inpaint the transition region for blended layout by the semantic layout generator~(See Sec.~\ref{SLG}).
(iii)~Sample a random noise $z_T\sim\mathcal{N}(0,\mathbf{I})$, then mix with $z_T^H$ and $z_T^B$ which are sampled from the forward noising process~(See Sec.~\ref{SGLDM}). Same for the following denoising steps.
(iv)~Condition the mixed noise by concatenating with the semantic latent representation $s$ at each denoising step.
(v)~Denoise from $z_T$ to $z_0$ and decode to $\tilde{x}$.

\begin{figure}[t]
  \centering
  \includegraphics[width=1\linewidth]{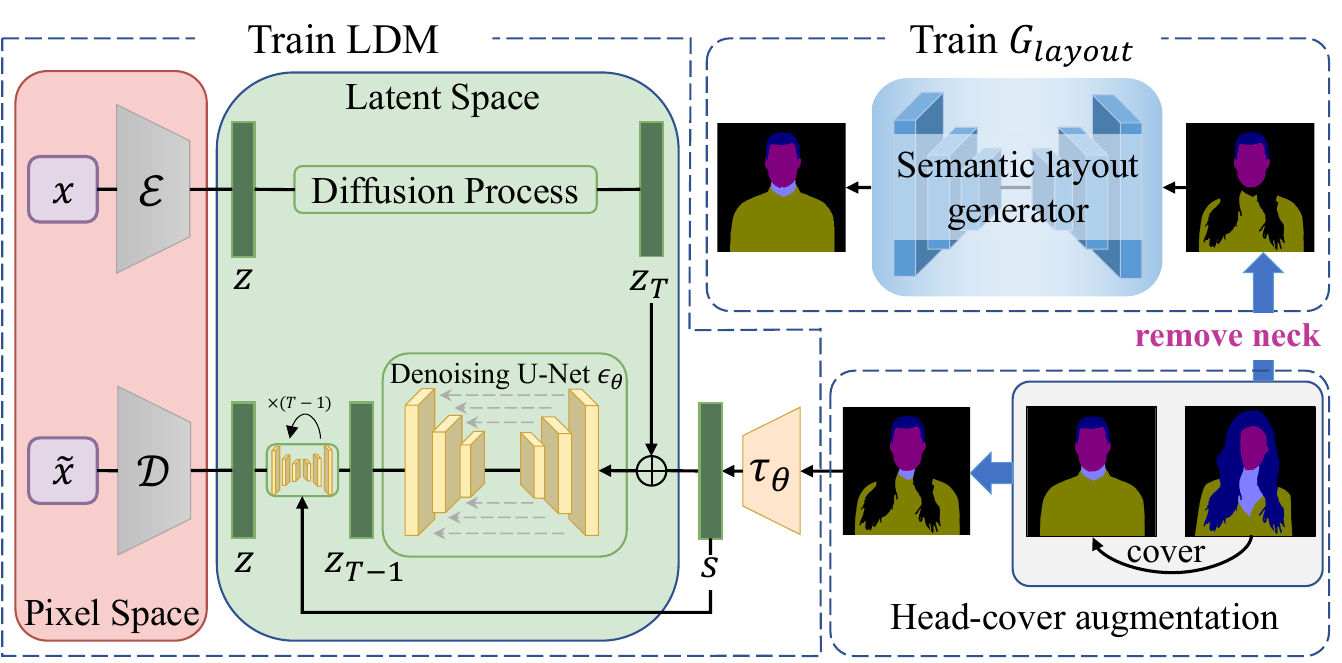}
  \caption{The training process. We train LDM and $G_{layout}$ separately with the head-cover augmentation for semantic calibration. Specifically, we remove the neck of input layout for $G_{layout}$.}
  \label{fig:train_process}
\end{figure}

\subsection{Semantic-Mixing LDM}
\label{SGLDM}
Latent Diffusion Model (LDM)~\cite{rombach2022high} can be trained to generate an image with the semantic layout as condition guidance.
As shown in Fig.~\ref{fig:train_process}, LDM consists of three components: a pretrained autoencoder~($\mathcal{E}$, $\mathcal{D}$)~\cite{esser2021taming}, a denoising U-Net $\epsilon_\theta$ and a condition encoder $\tau_\theta$.
More specifically, the encoder $\mathcal{E}$ can encode a half-body image $x$ to a latent code $z$~(i.e., $z=\mathcal{E}(x)$). The decoder $\mathcal{D}$ can reconstruct the half-body image from the latent code $z$~(i.e., $\tilde{x}=\mathcal{D}(z)$). With the high-quality reconstruction, the diffusion process can work in the low-dimensional latent space. $z_t$ can be directly sampled by $z_t = \sqrt{\bar{\alpha}_{t}}z_0+\sqrt{1-\bar{\alpha}_t}\epsilon$ as mentioned in Fig.~\ref{forward_noise}. The condition encoder $\tau_\theta$ encodes the layout $l$ to a latent representation $s$ as semantic guidance which is then concatenated with $z_t$ as the input of $\epsilon_\theta$ at each denoising step. Benefit from the spatial-level inductive biases from ($\mathcal{E}$, $\mathcal{D}$), the underlying denoising U-Net $\epsilon_\theta$ can be constructed with 2D convolution layers. And $\epsilon_\theta$ will further concentrate on the low-dimensional spatial-level representation in latent space efficiently, which is optimized by the reweighted variant of the variational lower bound:
\begin{equation}
\label{L_ldm}
     \mathcal{L}_{LDM} = \mathbb{E}_{z,s,\epsilon\sim\mathcal{N}(0,\mathbf{I}),t}\left[\|\epsilon-\epsilon_\theta(z_t,t,s)\|^2_2\right]
\end{equation}
where $\epsilon_\theta$ is trained to predict the noise $\epsilon$ contained in the input $z_t$ at any time $t$ under the semantic guidance $s$. When the $\mathcal{L}_{LDM}$ converges, iteratively denoising a $z_T\sim\mathcal{N}(0,\mathbf{I})$ to $z_0$~(See Fig.~\ref{ddim_denoise}) under the semantic guidance and then decoding $z_0$ can obtain a generated half-body image.

Head swapping expects to preserve the head of $x_1$ and the body of $x_2$, while ensuring the realistic transition and background regions. We can directly mix the latent representations $z_0^H$ and $z_0^B$ with corresponding masks~($m^H,m^B$) for head and body. But we cannot apply neck and background regions of either $x_1$ or $x_2$ to the head swapping results, because the neck size needs to fit the head and body, and the background region needs to consider the spatial occupation of the transition region. Therefore, we expect to expand or shrink adaptively the incomplete transition and background region. With the proposed semantic layout generator~(See Sec.~\ref{SLG}), we can obtain an inpainted semantic layout as condition guidance which provides plausible semantic information to guide denoising.

\begin{figure}[t]
  \centering
  \includegraphics[width=1\linewidth]{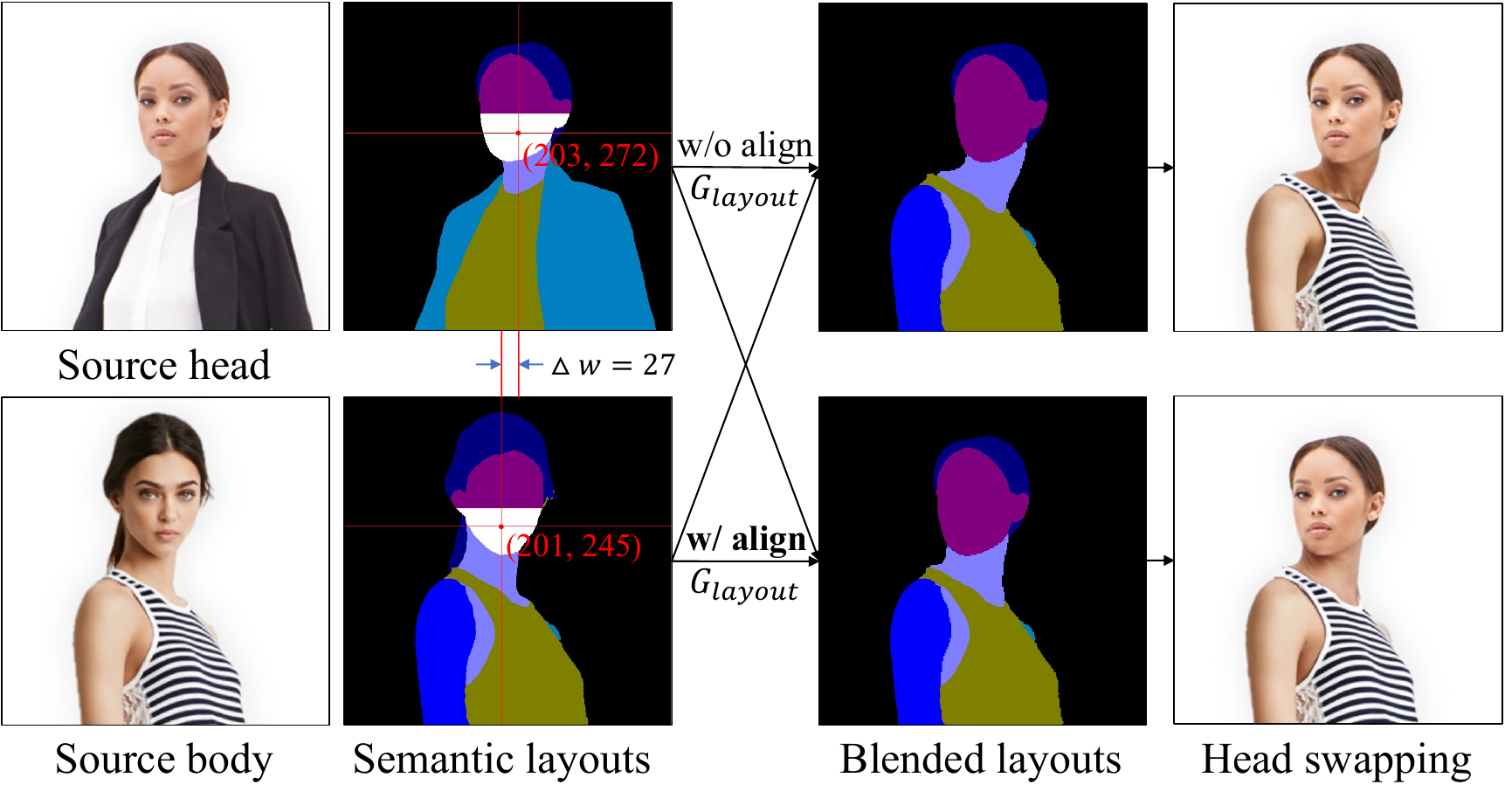}
  \caption{Neck alignment trick. We measure the horizontal deviation $\Delta w$ to align the upper boundary of neck from source head to source body, which makes the head swapping result more realistic.}
  \label{fig:neck_align}
\end{figure}

Inspired by the recent text-driven diffusion~\cite{avrahami2022blended_0,avrahami2022blended_1}, we design a progressive fusion strategy to implement head swapping with LDM. More specifically, we first sample a random $z_T\sim\mathcal{N}(0,\mathbf{I})$ and obtain $z_T^H$ and $z_T^B$ by forward noising process Fig.~\ref{forward_noise}, and they can be considered as noises from a image manifold of the same noising level. Then we mix these noises with corresponding masks, and the mixing at any time $t$ can be expressed as:
$\hat{z_t}=z_t^H\odot m^H + z_t^B\odot m^B + z_t\odot m^r$, where $m^r=1-m^H-m^B$ denotes the rest region.
Though the mixed noise $\hat{z_T}$ might deviate from this manifold, the next denoising step will fuse the non-unified regions in $\hat{z_T}$ and further land the output $z_{T-1}$ to the manifold at $T-1$ noising level. During the iteratively progressive fusion process, the regions $m^H$ and $m^B$ in $z_t$ are derived from forward noising process, while providing fundamental reference for generating $m^r$ in $z_t$. Under the semantic guidance, the region $m^r$ in $z_t$ will harmonize the boundaries to match $m^H$ and $m^B$ in $z_t$. In the end, we can obtain the mixed $z_0$ which appears to be a unity and decode to a seamless head swapping image $\tilde{x}$.

\subsection{Semantic Calibration}
Considering the possible errors in the blended semantic layout, (e.g., the hair covers the neck and body, annotation error), we propose a semantic calibration strategy to adaptively inpaint incomplete region and desensitize to semantic noise.
More specifically, we design an effective head-cover augmentation for training the LDM and semantic layout generator separately. As shown in Fig.~\ref{fig:train_process}, we randomly sample two half-body semantic layouts ($l_1,l_2$) from the training dataset, and use the head region of $l_2$ to cover the neck and body regions of $l_1$. And the covered region will be replaced with the background class. Since the randomly sampled ($l_1,l_2$) possess different scales on the head, neck and body regions, $l_2$ may be unchanged or covered by a small/large part in head and neck regions. Therefore, the multi-scale head-cover augmentation ensures the diversity for training and simulates as many cases as possible for head swapping. The proposed augmentation can effectively enable the semantic layout generator to inpaint and calibrate the incomplete layout, which can be used for coarse-grained head swapping at semantic level. Besides, it also endows our semantic-mixing LDM the semantic calibration capability to conduct a fine-grained head swapping even with a low-quality condition.

\subsection{Semantic Layout Generator}
\label{SLG}
To provide a plausible semantic guidance for head swapping with semantic-mixing LDM, we design a semantic layout generator $G_{layout}$ with a nested U-Net architecture~\cite{qin2020u2} trained in a self-supervised manner. More specifically, we introduce the proposed head-cover augmentation and further remove the neck region of input semantic layout $l$. To focus on the transition regions and leave the rest untouched, we employ the idea of focus map~\cite{nizan2020breaking} to add an extra output channel $m_{focus}$ for $G_{layout}(l)$. The final output $\tilde{l}$ is obtained by: $\tilde{l} = m_{focus}\odot \hat{l} + (1-m_{focus})\odot l$, where the $\hat{l}$ denotes the rest channels of $G_{layout}(l)$. Therefore, we incentivize $G_{layout}$ to inpaint the transition region adaptively by a pixel-wise cross-entropy loss and a LSGAN loss~\cite{mao2017least}:
\begin{equation}
\label{L_layout}
     \mathcal{L}_{layout} = \lambda_1\mathcal{L}_{CE} + \lambda_2\mathcal{L}_{GAN}
\end{equation}
where $\lambda_1$ and $\lambda_2$ are trade-off parameters. Since the argmax function is non-differentiable, we employ the Gumbel-softmax reparameterization trick \cite{jang2016categorical,li2021toward} to discretize the generated semantic layouts which allows the gradient to flow from the discriminator to $G_{layout}$. 
Besides, the generated semantic layouts are vulnerable to be discriminated as fake in the beginning of training, and the discretization is beneficial to avoid this situation.

When blending two semantic layouts ($l_1,l_2$) with the head mask $m^H$ and body mask $m^B$ directly, $G_{layout}$ can inpaint and calibrate the incomplete transition region of the blended layout for coarse-grained head swapping, as shown in Fig.~\ref{fig:train_process}. Based on the plausible semantic guidance provided by $G_{layout}$, semantic-mixing LDM will further calibrate the boundary pixels adaptively at each denoising process to conduct a fine-grained head swapping. It should be noted that without paired head swapping dataset, we have solved such a difficult problem with two self-supervised models, i.e., LDM and $G_{layout}$.

\begin{figure*}[t!]
  \centering
  \includegraphics[width=1\linewidth]{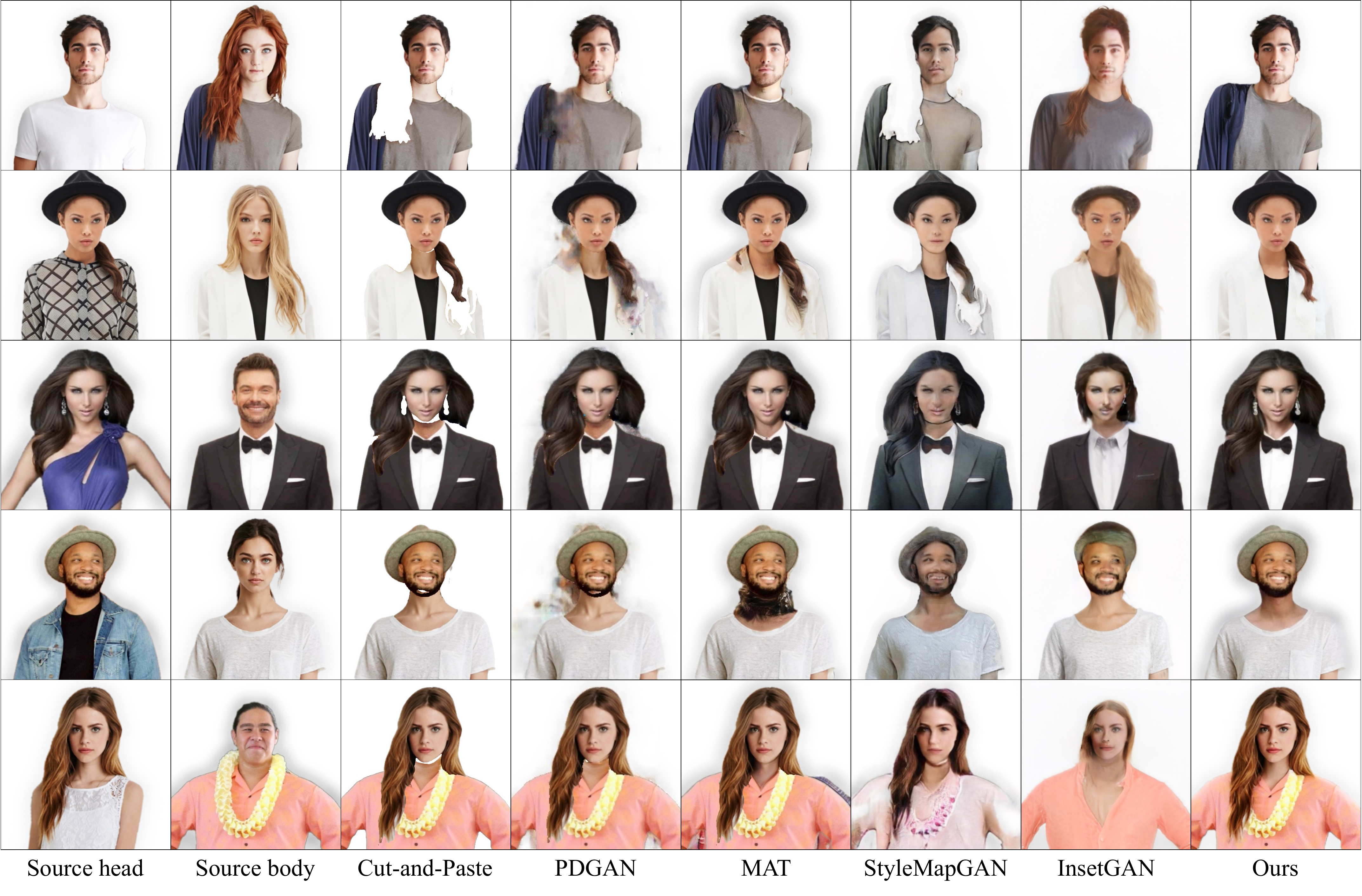}
  \caption{We present the qualitative comparisons with Cut-and-Paste, PDGAN~\cite{liu2021pd}, MAT~\cite{li2022mat}, StyleMapGAN~\cite{kim2021exploiting} and InsetGAN~\cite{fruhstuck2022insetgan} on our half-body SHHQ256 dataset. Our head swapping results show overall superior quality with flawless preservation and seamless transition for the source head and source body. The qualitative results without the proposed neck alignment trick can be found in the supplementary material.}
  \label{fig:comparsion}
\end{figure*}

\subsection{Neck Alignment Trick}
Face alignment~\cite{kazemi2014one} can normalize the size of heads in dataset to a same level and automatically align faces to a same position.
However, if the face orientations of two face-aligned images~($x_1,x_2$) are different, there may be a horizontal deviation between their necks as shown in Fig.~\ref{fig:neck_align}, which will affect the realism of head swapping results.
We observe problems with the neck regions which are difficult to distinguish from the chest skin and often covered by clothes, so we cannot directly address the neck deviation with the neck regions.
Fortunately, we find that the lower face~(i.e., the face region below the nose landmark) is hardly covered and its center coordinate can roughly indicate the whole head position.
Thus we measure the horizontal deviation $\Delta w$ between two center coordinates in the layouts~($l_1,l_2$) and move the source head to align to the source body, which is equivalent to aligning the upper boundary of the neck. This trick solves the neck alignment problem without training parameters and enables the downstream model to generate more realistic head swapping results. In addition, due to the rotatability of the human head, the head swapping results can enhance geometric realism with this trick even if the face orientation of the source head and source body images are different as shown in Fig.~\ref{fig:neck_align}.

\section{Experiment}
\label{sec:experiment}

\subsection{Experimental Setting}
\label{Exp_Set}
\noindent\textbf{Dataset.}
The Stylish-Humans-HQ Dataset (SHHQ-1.0)~\cite{fu2022stylegan} consists of 39,942 full-body images which are aligned with the body center. We reprocess the SHHQ-1.0 dataset with a face alignment technique~\cite{kazemi2014one} and crop out the half-body images as our half-body SHHQ dataset. In addition, we use a SOTA human parsing method SCHP~\cite{li2020self} to obtain the semantic layouts of half-body images. We randomly select 35,942 half-body images as the training set, and use the remaining 4,000 images as the testing set where the source head images and source body images are each half. And we conduct experiments on the half-body SHHQ256 and half-body SHHQ512 datasets.

\begin{table*}[!t]
\centering
\setlength{\tabcolsep}{7pt}
\caption{Quantitative comparisons with baselines on our half-body SHHQ256 dataset. $\downarrow$ indicates that lower is better, while $\uparrow$ indicates higher is better. The $1^{st}/2^{nd}/3^{rd}$ best results of competing methods are indicated in \textbf{\textcolor{red}{red}}/\textbf{\textcolor{blue}{blue}}/\textbf{\textcolor{black}{black}}. We also provide the comparison results on the half-body SHHQ512 in our supplementary material.
 }
\label{tab:comparsion}
\begin{tabular}{lcccccccc}
\toprule
\multirow{2}{*}{\textbf{Methods}}      & \multirow{2}{*}{IDs$\uparrow$} & \multicolumn{2}{c}{Head preservation}      & \multicolumn{2}{c}{Body preservation} & \multirow{2}{*}{FID$\downarrow$} & \multirow{2}{*}{Mask-FID$\downarrow$} & \multirow{2}{*}{Focal-FID$\downarrow$}  \\
&                                 & SSIM$\uparrow$              & LPIPS$\downarrow$  & SSIM$\uparrow$              & LPIPS$\downarrow$             &                             &                            &                      \\
\midrule

Cut-and-Paste                          & $-$                               & $-$                 & $-$      & $-$                 & $-$           & 26.22         & $-$                      & 31.48                                 \\
+ Neck Alignment Trick& $-$                               & $-$                 & $-$      & $-$                 & $-$              & 26.17        & $-$                        &  31.18                                \\
\hdashline\specialrule{0em}{1pt}{1pt}
PDGAN                                 & 0.9826                               & 0.9906             & 0.0095   &  \textbf{\textcolor{blue}{0.9702}}          & \textbf{\textcolor{blue}{0.0413}}         & 23.83     & 56.68                       & 37.98                                    \\
+ Neck Alignment Trick           & \textbf{\textcolor{blue}{0.9885}}                               & \textbf{\textcolor{blue}{0.9941}}           & \textbf{\textcolor{blue}{0.0081}}   & 0.9697             & 0.0422          & \textbf{23.72}    & 57.15                       & 38.66                                    \\
\hdashline\specialrule{0em}{1pt}{1pt}
MAT                                   & 0.9883                               & 0.9968            & 0.0008 &   \textbf{\textcolor{red}{0.9719}}          & \textbf{\textcolor{red}{0.0372}}        & 16.64      & 35.05                       & 19.51                                    \\
+ Neck Alignment Trick             & \textbf{\textcolor{red}{0.9899}}                               & \textbf{\textcolor{red}{0.9979}}            & \textbf{\textcolor{red}{0.0007}} &  0.9713           & 0.0383        & \textbf{\textcolor{blue}{16.11}}      & \textbf{33.28}                       & \textbf{\textcolor{blue}{18.74}}                                    \\
\hdashline\specialrule{0em}{1pt}{1pt}
StyleMapGAN & 0.7553 & 0.8956 & 0.0638 &  0.8170  & 0.1295 & 32.25 & 25.51 & 32.88 \\
+ Neck Alignment Trick & 0.7567  & 0.8992 & 0.0606 & 0.8166  & 0.1278 & 31.51 & \textbf{\textcolor{blue}{24.44}} & 31.94                \\

\hdashline\specialrule{0em}{1pt}{3pt}
InsetGAN                           &  0.8235 & 0.8670 & 0.0936 & 0.8085 & 0.1157 & 28.18 & 47.91 & \textbf{25.58} \\

+ Neck Alignment Trick    &  0.8227 & 0.8673 & 0.0962 & 0.8097 & 0.1144 & 28.39 & 48.46 & 25.78                 \\
\midrule
Ours                                   & 0.9783                               & 0.9686            & 0.0237 & 0.9308            & 0.0518           & 11.45        & 19.86                       & 12.34                               \\
+ Neck Alignment            & \textbf{0.9812}                  & \textbf{0.9689}            & \textbf{0.0233} & \textbf{0.9310}             & \textbf{0.0517}           & \textbf{\textcolor{red}{11.24}}    & \textbf{\textcolor{red}{18.57}}                       & \textbf{\textcolor{red}{11.80}}                         \\

\bottomrule   
\end{tabular}
\end{table*}

\noindent\textbf{Implementation Details.} We choose the downsampling factor $f=4$ for the latent code $z$ and the semantic representation $s$ which is the best setting in LDM~\cite{rombach2022high}. We adopt an Adam optimizer~\cite{kingma2014adam} with momentum parameters $\beta_1= 0.5$ and $\beta_2 = 0.999$ to optimize all models. The trade-off parameters $\lambda_1$ and $\lambda_2$ for training the semantic layout generator $G_{layout}$ are set to $1$ and $0.2$. The performance evaluation of $G_{layout}$ will be discussed in the supplementary material.
All the experiments are carried out on a server with 8 Nvidia V100 GPUs.

\noindent\textbf{Baselines.} To the best of our knowledge, there is no available image-based head-swapping method. Therefore, we use four recent methods designed for similar tasks to implement this task. We also provide the results of directly cutting a source head and another source body$+$neck, and then pasting on a canvas~(Cut-and-Paste).
We introduce \textit{two SOTA inpainting methods} PDGAN~\cite{liu2021pd} and MAT~\cite{li2022mat} trained by the proposed head-cover augmentation and removing the neck region. And when testing, head swapping can be implemented by inpainting the results of Cut-and-Paste without neck.
We also compare with \textit{two SOTA image editing methods}: The StyleMapGAN~\cite{kim2021exploiting} is designed for face editing with spatial latent code (downsampling factor $f=32$), we fairly set $f=4$ for training on our half-body SHHQ dataset as we do. With the well-trained StyleMapGAN, we can implement head swapping by semantic manipulation. Based on a well-trained half-body StyleGAN2~\cite{karras2020analyzing} and a face StyleGAN2, InsetGAN~\cite{fruhstuck2022insetgan} can swap generated face by a multi-optimization process on latent codes. We obtain the training set for the face StyleGAN2 by aligning and cropping the half-body SHHQ like FFHQ~\cite{karras2019style}. To achieve head swapping for real images, we train an e4e encoder~\cite{tov2021designing} with the half-body StyleGAN2 to obtain the latent code of half-body image. And we use the StyleGAN2 Projection~\cite{karras2020analyzing} to project face image to a latent code with the face StyleGAN2. With the latent codes of source head image and source body image, head swapping can be achieved by optimization of InsetGAN.

\noindent\textbf{Evaluation Metrics.} To evaluate the head swapping results, we adopt four common quantitative evaluation metrics: 
$\diamond$ \textbf{Identity similarity~(IDs)} measures the average cosine similarity between face embeddings extracted by ArcFace~\cite{deng2019arcface}.
$\diamond$ \textbf{SSIM}~\cite{wang2004image} is a perceptual metric which measures structural similarity. $\diamond$ \textbf{LPIPS}~\cite{zhang2018perceptual} is based on the AlexNet~\cite{krizhevsky2012imagenet}, which have been demonstrated consistency with human perception. 
Since we expect that the source head and source body can be reconstructed well, we can calculate the SSIM and LPIPS only on the head and body regions respectively.
$\diamond$ \textbf{Fréchet Inception Distance (FID)}~\cite{lucic2017gans}: FID measures the Earth-Mover Distance (EMD) between the feature distributions of generated images and real images.

Though FID does not need paired ground truths to evaluate the head swapping results, it considers the whole image and does not focus on the edited region. Therefore, we propose two tailor-designed improvements of FID to further compare with baselines: $\diamond$ \textbf{Mask-FID}: We mask the head and body regions of head swapping results and test set to expose the inpainting regions and then calculate the FID.
$\diamond$ \textbf{Focal-FID}: Since the generated transition regions are mainly in the center of the half-body images, We crop out the middle $1/2$ region horizontally and vertically for all head swapping results and test set to calculate the FID.

\begin{table*}[!t]
\centering
\setlength{\tabcolsep}{9pt}
\caption{Quantitative comparisons with baselines on our half-body SHHQ512 dataset. $\downarrow$ indicates that lower is better, while $\uparrow$ indicates higher is better. The $1^{st}/2^{nd}/3^{rd}$ best results of competing methods are indicated in \textbf{\textcolor{red}{red}}/\textbf{\textcolor{blue}{blue}}/\textbf{\textcolor{black}{black}}.
 }
\label{tab:comparsion512}
\begin{tabular}{lcccccccc}
\toprule
\multirow{2}{*}{\textbf{Methods}}      & \multirow{2}{*}{IDs$\uparrow$} & \multicolumn{2}{c}{Head preservation}      & \multicolumn{2}{c}{Body preservation} & \multirow{2}{*}{FID$\downarrow$} & \multirow{2}{*}{Mask-FID$\downarrow$} & \multirow{2}{*}{Focal-FID$\downarrow$}  \\
&                                 & SSIM$\uparrow$              & LPIPS$\downarrow$  & SSIM$\uparrow$              & LPIPS$\downarrow$             &                             &                            &                      \\
\midrule

Cut-and-Paste                          & $-$                               & $-$                 & $-$      & $-$                 & $-$           & \textbf{26.22}         & $-$                      & 31.48                                 \\

PDGAN                                 & \textbf{0.9828}                               & \textbf{\textcolor{blue}{0.9957}}             & \textbf{0.0236}   &  \textbf{\textcolor{blue}{0.9546}}          & \textbf{0.0605}         & 38.05     & 77.43                       & 37.02                                   \\

MAT                                   & \textbf{\textcolor{red}{0.9917}}                               & \textbf{\textcolor{red}{0.9959}}            & \textbf{\textcolor{red}{0.0006}} &   \textbf{\textcolor{red}{0.9710}}          & \textbf{\textcolor{red}{0.0299}}        & \textbf{\textcolor{blue}{12.84}}      & \textbf{\textcolor{blue}{33.96}}                       & \textbf{\textcolor{blue}{18.53}}                                  \\

StyleMapGAN & 0.8368 & 0.9145 & 0.0416 &  0.8332  & 0.1068 & 30.38 & \textbf{35.05} & 35.71 \\

InsetGAN                            &  0.8247 & 0.8712 & 0.0963 & 0.8096 & 0.1153 & 29.96 & 49.88 & \textbf{27.98} \\

\midrule

Ours            & \textbf{\textcolor{blue}{0.9913}}                  & \textbf{0.9813}            & \textbf{\textcolor{blue}{0.0125}} & \textbf{0.9545}             & \textcolor{blue}{\textbf{0.0354}}           & \textbf{\textcolor{red}{10.73}}    & \textbf{\textcolor{red}{21.26}}                       & \textbf{\textcolor{red}{11.42}}                         \\

\bottomrule   
\end{tabular}
\end{table*}

\begin{table}[t]
\centering
\setlength{\tabcolsep}{2pt}
\caption{Ablation study presents the quantitative scores of introducing our proposed head-cover augmentation and semantic layout generator $G_{layout}$ separately and jointly based on LDM.}
\label{tab:ablation_fid}
\begin{tabular}{lccc}
\toprule
 & FID$\downarrow$   & Mask-FID$\downarrow$  & Focal-FID$\downarrow$  \\
 \midrule
LDM   & 35.36 & 57.85      & 55.26     \\
\hdashline\specialrule{0em}{1pt}{1pt}
+ Head-cover & 14.43 & 26.51       & 16.82     \\
\hdashline\specialrule{0em}{1pt}{1pt}
+ $G_{layout}$  & 12.24 & 20.74 & 14.20      \\
\hdashline\specialrule{0em}{1pt}{2pt}
+ Head-cover + $G_{layout}$ & \textbf{11.24} & \textbf{18.57} & \textbf{11.80}\\
\bottomrule  
\end{tabular}
\end{table}

\subsection{Comparison}
\subsubsection{Qualitative comparison.}
As shown in Fig.~\ref{fig:comparsion}, we show all head swapping results with our neck alignment trick, which can effectively enhance geometric realism. We also present more qualitative comparisons Fig.~\ref{fig:a4}.

The Cut-and-Paste indicates that the head swapping task needs to inpaint a transition region as seamless connection, while preserving the appearance of the source head and source body as much as possible. The PDGAN shows moderate performance for inpainting neck regions, but fails to inpaint the covered regions and brings obvious noise and artifacts. The transformer-based MAT produces better results, but the inpainting regions do not harmonize with the surroundings well. Without semantic guidance, the inpainting results of MAT are uncontrollable and even expands outward from the body region with undesired texture. The StyleMapGAN blends the spatial latent codes of source head and source body and produces decent performance for the neck regions. However, even though we have increased the dimension of the spatial latent code, StyleMapGAN still fails to reconstruct the half-body image. The optimization-based InsetGAN also suffers from bad preservation for head and body. Fortunately, our framework can inpaint the transition region seamlessly while preserving the source head and source body with high-quality reconstruction.

\begin{figure}[t!]
  \centering
  \includegraphics[width=1\linewidth]{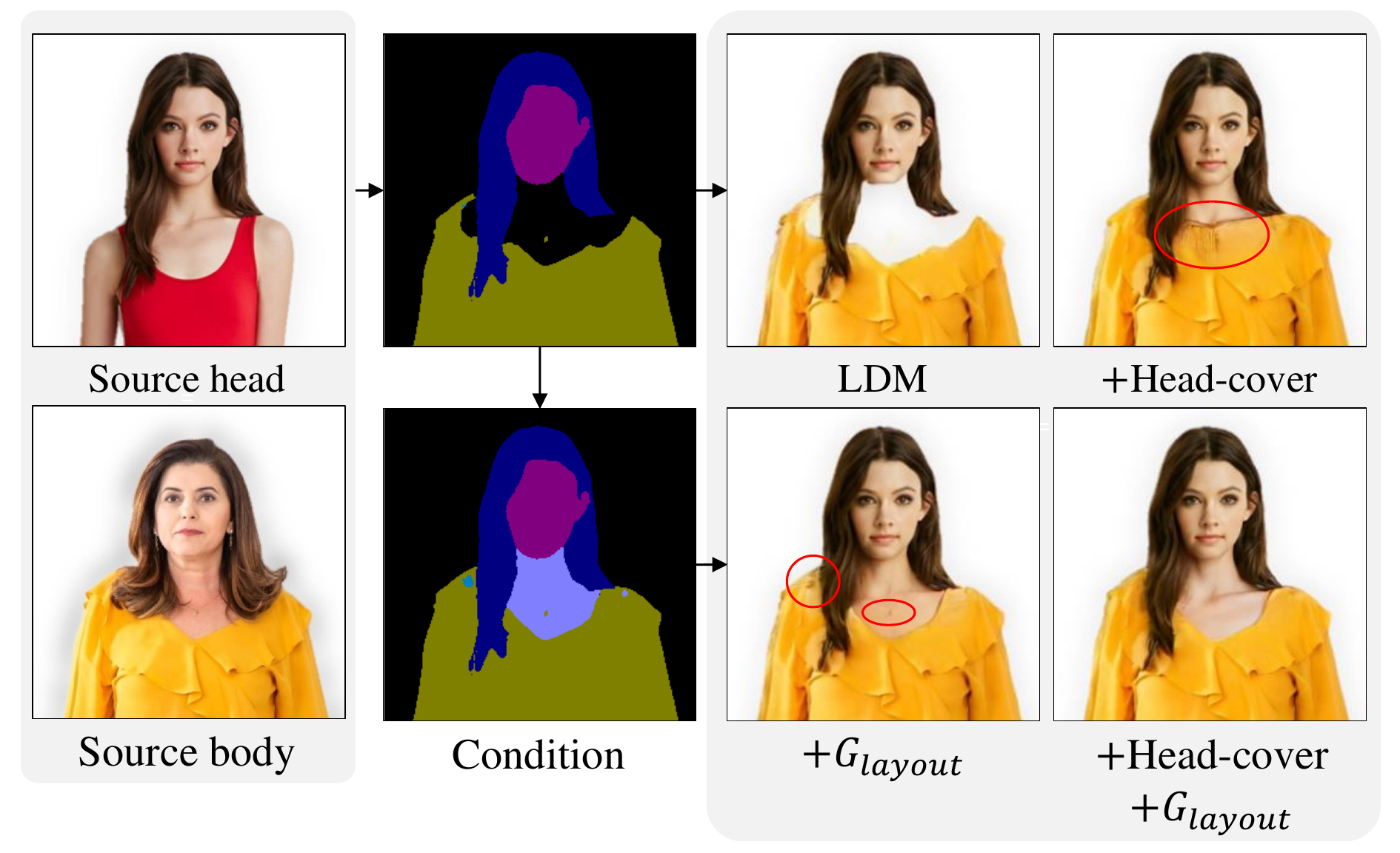}
  \caption{Ablation study. The LDM and ``+Head-cover" setting are conditioned by incomplete layout. And the ``+$G_{layout}$" and the joint setting are conditioned by blended layout.}
  \label{fig:ablation}
\end{figure}

\subsubsection{Quantitative comparison.}
\noindent\textbf{On half-body SHHQ256 dataset.}
As shown in Table~\ref{tab:comparsion}, we not only present the quantitative results of all baselines, but also fairly provide the results with our proposed neck alignment trick. To evaluate the head preservation and body preservation, we measure SSIM and LPIPS with the head masks and body masks separately. To measure the ``+ Neck Alignment Trick" settings for all methods, we pan the ground-truth heads to align with the corresponding head swapping results to calculate quantitative scores. Since the Cut-and-Paste directly cuts the source head and the source body$+$neck, and then paste on a canvas, we do not compare its IDs, preservation scores and Mask-FID. Instead, its FID and Focal-FID can be considered as a baseline.

The results of Cut-and-Paste are incomplete at the transition region, so it obtains a poor FID~(26.17) which is further exposed by its result 31.18 on our proposed Focal-FID. The inpainting methods PDGAN and MAT focus on filling in the incomplete transition regions, and hardly change the source head and source body which are expected to be preserved. Therefore, they achieve the best scores on IDs, SSIM, and LPIPS. However, FID indicates that their head swapping results are not optimal. The proposed Mask-FID and Focal-FID can diminish the disturbance of the preserved source head and source body, and as it happens, PDGAN obtains the worst Mask-FID 56.68 and Focal-FID 37.98. MAT only makes decent Mask-FID 33.28 and Focal-FID 18.74. The latent codes of the image editing methods StyleMapGAN and InsetGAN are difficult to preserve the identity and texture for the half-body images, thus only achieving decent results in terms of IDs and keeping scores. The optimization-based InsetGAN leads to more harmonious boundaries of the head-swap results by optimizing the latent codes of face image and half-body image, hence it make a good Focal-FID 25.58. But the Mask-FID specifically measures the quality of the inpainting regions and reveals the weakness of InsetGAN's poor generation quality. In contrast, our method not only makes satisfied source head and source body preservation, but also achieves optimal image quality, and outperforms the $2^{nd}$ by 4.87/5.87/6.94 on FID/Mask-FID/Focal-FID respectively.

Obviously, the ``+ Neck Alignment Trick" settings improve all FIDs for all methods except InsetGAN, which is consistent with our expectation. The reason for FIDs boost is that our trick enables the downstream model to generate more realistic head swapping results.
The reason for the discord of InsetGAN with this trick is that the well-trained face StyleGAN2 is sensitive to whether the face images are aligned or not. This trick moves the face's position so that it's difficult to invert the high-quality latent codes by the face StyleGAN2.

\begin{figure}[t]
  \centering
  \includegraphics[width=1\linewidth]{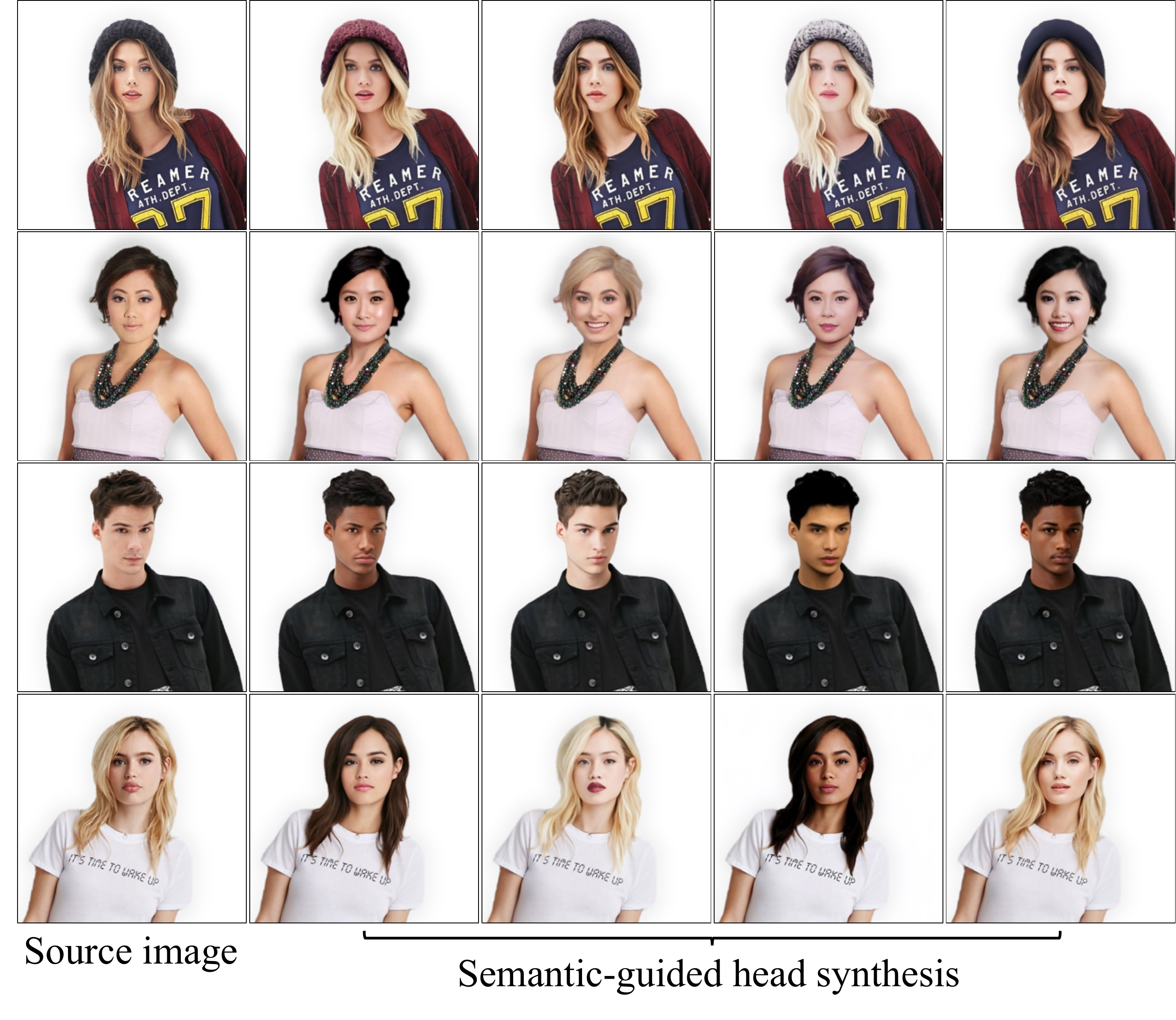}
  \caption{Results of semantic-guided head replacement. We can replace the head in a real image with fake, which can be sampled with diverse hat colors, hair colors, skin tones, identities and expressions under the semantic guidance.}
  \label{fig:random_head}
\end{figure}

\noindent\textbf{On half-body SHHQ512 dataset.}
We not only show the quantitative comparisons with the half-body SHHQ256 dataset in Table~\ref{tab:comparsion}, but also demonstrate our superiority on the half-body SHHQ512 dataset as shown in Fig.~\ref{tab:comparsion512}. All baselines except InsetGAN~\cite{fruhstuck2022insetgan} are conducted with our neck alignment trick. Our framework makes high-quality reconstruction which significantly surpasses the latent-space editing methods StyleMapGAN~\cite{kim2021exploiting} and InsetGAN~\cite{fruhstuck2022insetgan}. 
Besides, we also surpass the $2^{nd}$ by 2.11/12.7/7.11 on FIDs (i.e., FID, Mask-FID and Focal-FID), which are the key to compare with the head swapping results of competing methods. The outstanding quantitative comparisons are consistent with our superior qualitative results which demonstrates the effectiveness of our semantic-mixing LDM and the semantic calibration strategy.

\begin{figure}[t!]
  \centering
  \includegraphics[width=1\linewidth]{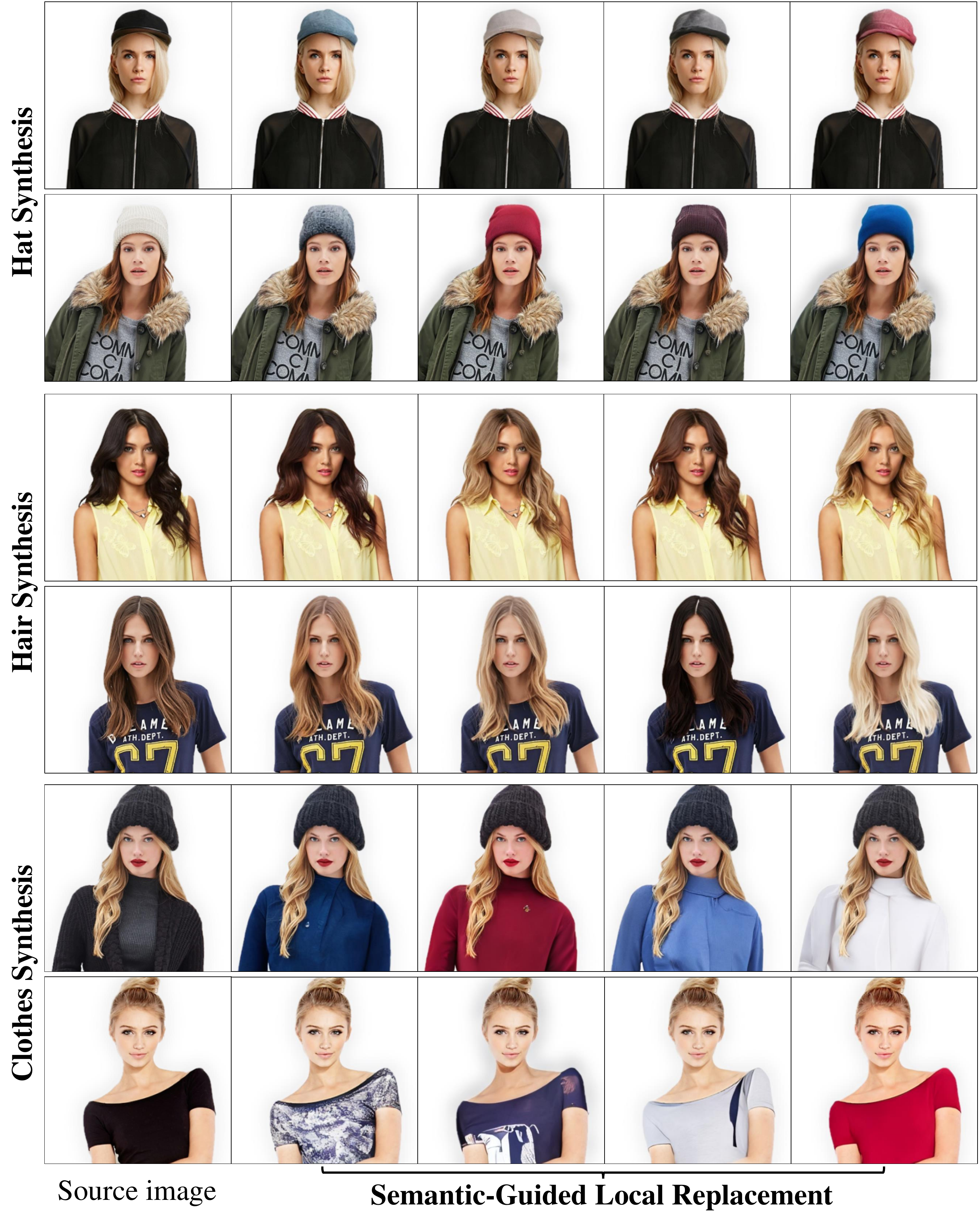}
  \caption{We present the semantic-guided local replacement on hat, hair and clothes regions of \textbf{real images} The replaced regions can be seamlessly stitched to the other regions.}
  \label{fig:random_locals}
\end{figure}

\begin{figure*}[t!]
  \centering
  \includegraphics[width=0.95\linewidth]{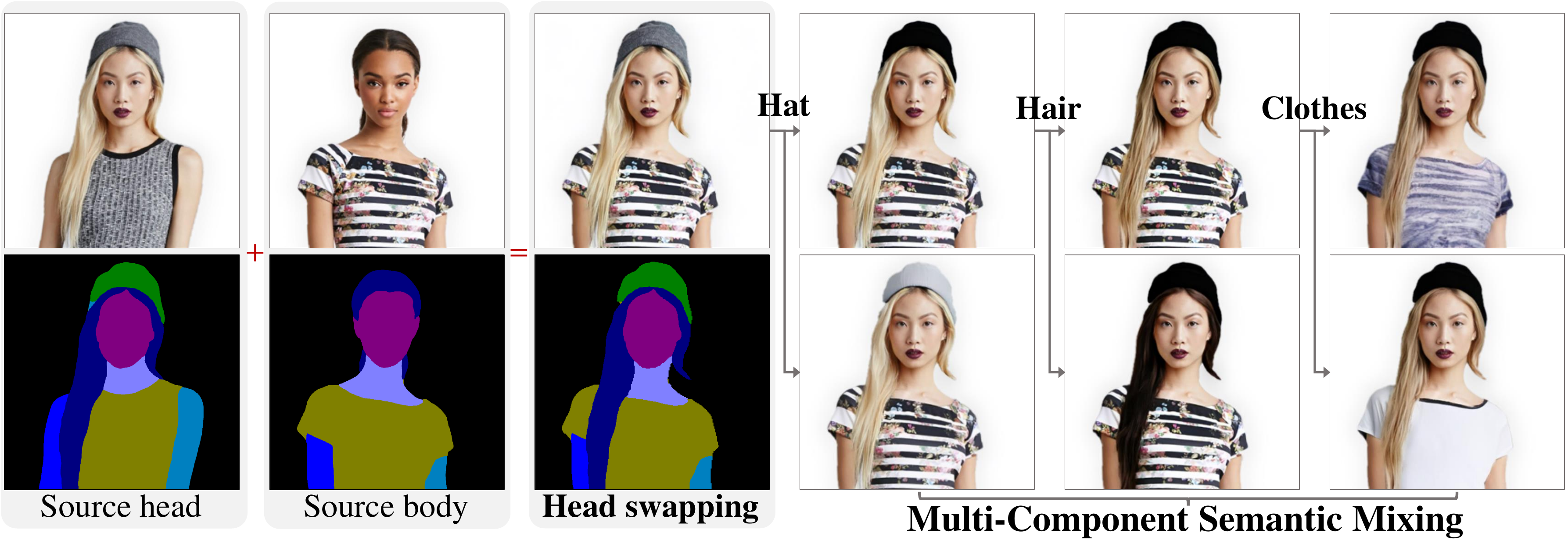}
  \caption{Multi-component semantic mixing~(hat, hair and clothes) on a \textbf{head swapping result}. Each component replacement can produce diverse and high-quality results.}
  \label{fig:multi_component}
\end{figure*}

\subsection{Ablation Study}
\label{section_ablation}
All ablation experiments are conducted with the neck alignment trick, which has been proven effective in Table~\ref{tab:comparsion}. All results are obtained by the semantic-mixing way as shown in Fig.~\ref{fig:infer_process}. We discuss qualitative and quantitative performance of the head-cover augmentation and $G_{layout}$ upon LDM respectively, and the superiority when working jointly. Since these settings hardly affect the head/body preservation, we only compare on FIDs (i.e., FID, Mask-FID and Focal-FID).

\begin{figure}[t!]
  \centering
  \includegraphics[width=1\linewidth]{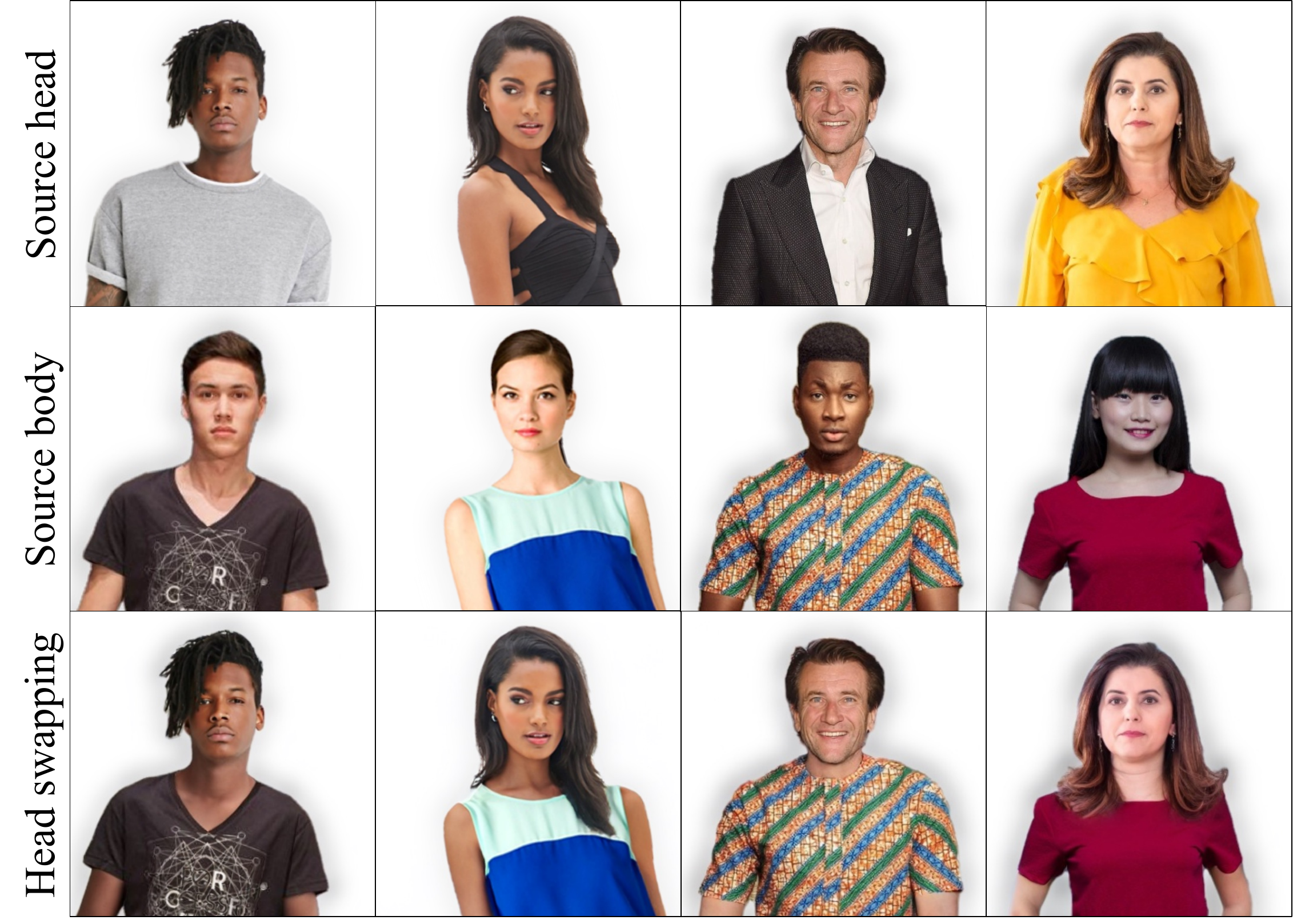}
  \caption{Cross-skin-tone head swapping. The skin tone of head swapping results is consistent with the source head.}
  \label{fig:skin_tone}
\end{figure}

As shown in Fig.~\ref{fig:ablation}, the naive LDM only can follow the condition to generate specified region. To spur the LDM to actively generate the transition regions under incomplete condition, we train the LDM with head-cover augmentation and removing the neck region. The ``+Head-cover" setting can inpaint the transition region with autonomous drawing and significantly improve the FIDs of the head swapping results as shown in Table~\ref{tab:ablation_fid}. But we expect to keep the body unchanged without extending more clothes and other regions. Therefore, we introduce the $G_{layout}$ to calibrate the 
coarse-grained condition for LDM and achieve better FIDs. $G_{layout}$ can inpaint the input layout well, but where a small annotation error may affect the subsequent results. So we combine the ``+Head-cover" setting and ``+$G_{layout}$" setting to implement the coarse-to-fine head swapping, where semantic-mixing LDM can further calibrate details and produce fine-grained results. The satisfied results in visual are consistent with the superior quantitative results.

\begin{figure}[t!]
  \centering
  \includegraphics[width=1\linewidth]{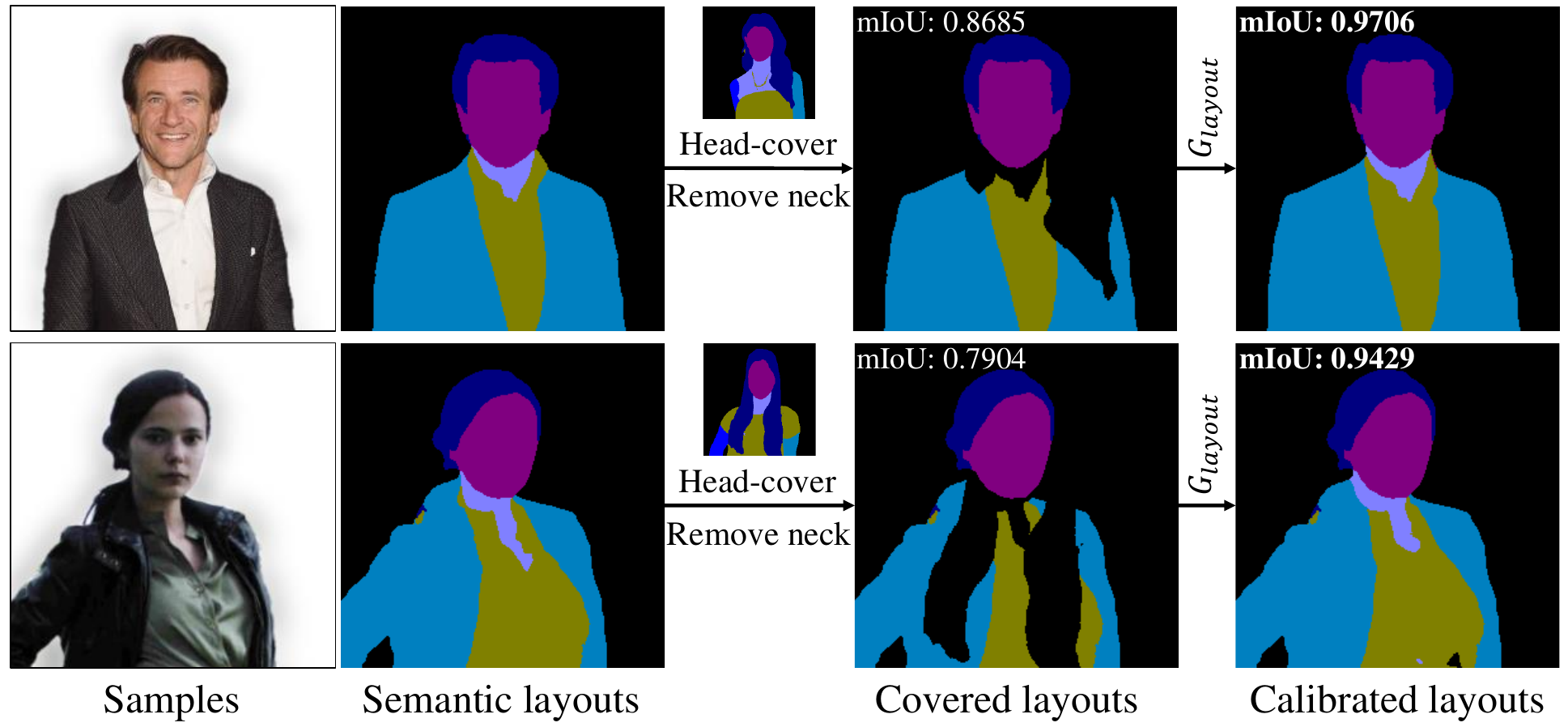}
  \caption{With the semantic calibration for layouts, the covered region can be inpainted by $G_{layout}$ in accordance with human perception and the mIoU is improved.}
  \label{fig:calibrate_layouts}
\end{figure}

\begin{figure*}[t!]
  \centering
  \includegraphics[width=1\linewidth]{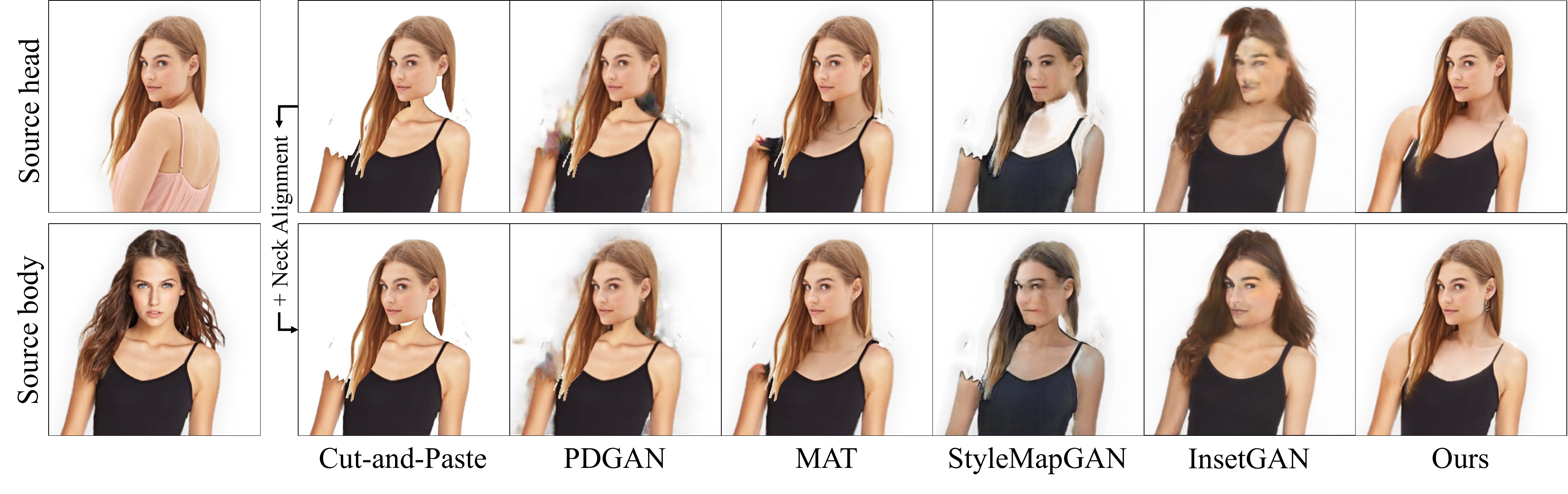}
  \caption{Effect of the proposed neck alignment trick. We present the qualitative comparison to show the visual effect of the trick. The first line and the second line are w/o the trick and w/ the trick respectively. Obviously, the proposed neck alignment trick makes the head swapping results of all baselines more realistic.}
  \label{fig:effect_neck_align}
\end{figure*}

\subsection{Head Replacement with Fake}
\label{sec:head_fake}
When user wants to replace the head in a real image $x$ with fake, only the $z_T$ of head and neck regions need to be sampled randomly from $\mathcal{N}(0,\mathbf{I})$ and then mixed with the preserved body region $z_t^B(t=T,\cdot\cdot\cdot,1)$ as the denoising steps in Fig.~\ref{fig:infer_process}. The condition is encoded by the semantic layout $l$ of $x$, so the layout of fake image will be consistent with $l$, as shown in Fig.~\ref{fig:random_head}. Under this setting, we can sample diverse hat colors, hair colors, skin tones, identities, and expressions with photo-realistic texture in the head region.

\section{Discussion}
\label{sec:discussion}

\subsection{Semantic-Guided Local Replacement}
In addition to Fig.~\ref{fig:random_head}, we further conduct the semantic-guided local replacement for hat, hair, and clothes regions as shown in Fig.~\ref{fig:random_locals}. As we introduced in Sec.~\ref{sec:head_fake}, we only sample the region that we expect to replace and make a seamless transition with the preserved regions by our semantic-mixing LDM under semantic guidance. We can sample diverse colors and textures for the replacement of hat, hair, and clothes regions. Furthermore, we achieve multi-component semantic mixing by conducting semantic-guided local replacement step by step as shown in Fig.~\ref{fig:multi_component}. The outstanding results demonstrate the superiority and versatility of our framework.

\subsection{Cross-skin-tone Head Swapping}
When there is a clear difference in skin tone between the source head and source body, we expect to resample the skin tone of body region to match source head out of the respect for source head. To this end, we sample the transition region and the regions of human limbs and blend them with the source head and source clothes by our semantic-mixing LDM. As shown in Fig.~\ref{fig:skin_tone}, the skin tone of head swapping results is consistent with the source head. This further proves the effectiveness of our head swapping framework.

\subsection{Effect of $G_{layout}$}
We evaluate the effect of $G_{layout}$ for semantic calibration on our half-body SHHQ256 test set. More specifically, we remove the neck region for each semantic layout in the test set and randomly introduce another layout in the test set to conduct the head-cover augmentation as shown in Fig.~\ref{fig:calibrate_layouts}. And we use $G_{layout}$ to inpaint the covered layout and the mIoU of covered layout is improved. With the semantic calibration, although the semantic layout cannot be reconstructed exactly, the covered region can be inpainted in accordance with human perception. Therefore, $G_{layout}$ can calibrate the blended layout for head swapping. And we experiment with 10 times random seeds on our test set and achieve $0.9135\pm0.0023$ performance on the mean Intersection over Union~(mIoU), where we achieves $0.9319\pm0.0016$ on IoU for the neck region. This demonstrates the effectiveness of our semantic calibration strategy, where $G_{layout}$ can provide plausible semantic layouts.

\subsection{Effect of Neck Alignment Trick}
We have demonstrated the proposed neck alignment trick will improve the performance of all competing methods except InsetGAN by the quantitative results in Table~\ref{tab:comparsion}. In addition, we further show the visual effect of the proposed trick in Fig.~\ref{fig:effect_neck_align}. This trick will move the source head to align to the head of source body, which assists the downstream models to produce more realistic head swapping results. Although face-aligned side-face images are often difficult to invert, our trick moves the entire head region to the region of interest for the face StyleGAN2~\cite{karras2020analyzing}, which allows InsetGAN to achieve better head swap results on the side-face images. Compared to these methods, our framework seamlessly stitches source head to source body and generates a flawless transition region while preserving high-quality reconstruction for the source head and source body. And when conducting with the proposed trick, we achieve more natural-looking head swapping result.

\section{Conclusion}
\label{sec:conclusion}
In this paper, we propose a new image-based head swapping framework which is implemented by a semantic-mixing LDM and a semantic layout generator. We train our framework with the proposed head-cover augmentation in a self-supervised manner for semantic calibration. And the proposed neck alignment trick will align the source head to a position where the downstream model can produce more geometric-realistic head swapping results. Furthermore, we construct a new image-based head swapping benchmark and propose two improvements of FID (i.e., Mask-FID and Focal-FID) to further compare with baselines.

\textbf{Broader Impact.} Although the face and head swapping technologies can bring great commercial value, it may be used for the unethical behaviors, such as identity forgery. To prevent the potential risks and promote the healthy development of AI, we will provide our head swapping results to the face/head forgery detection community.

\appendix

\onecolumn
\section*{\centering{HS-Diffusion: Semantic-Mixing Diffusion for Head Swapping (Appendix)}}

\begin{center}
\captionsetup{type=figure}
  \centering
  \includegraphics[width=1\linewidth]{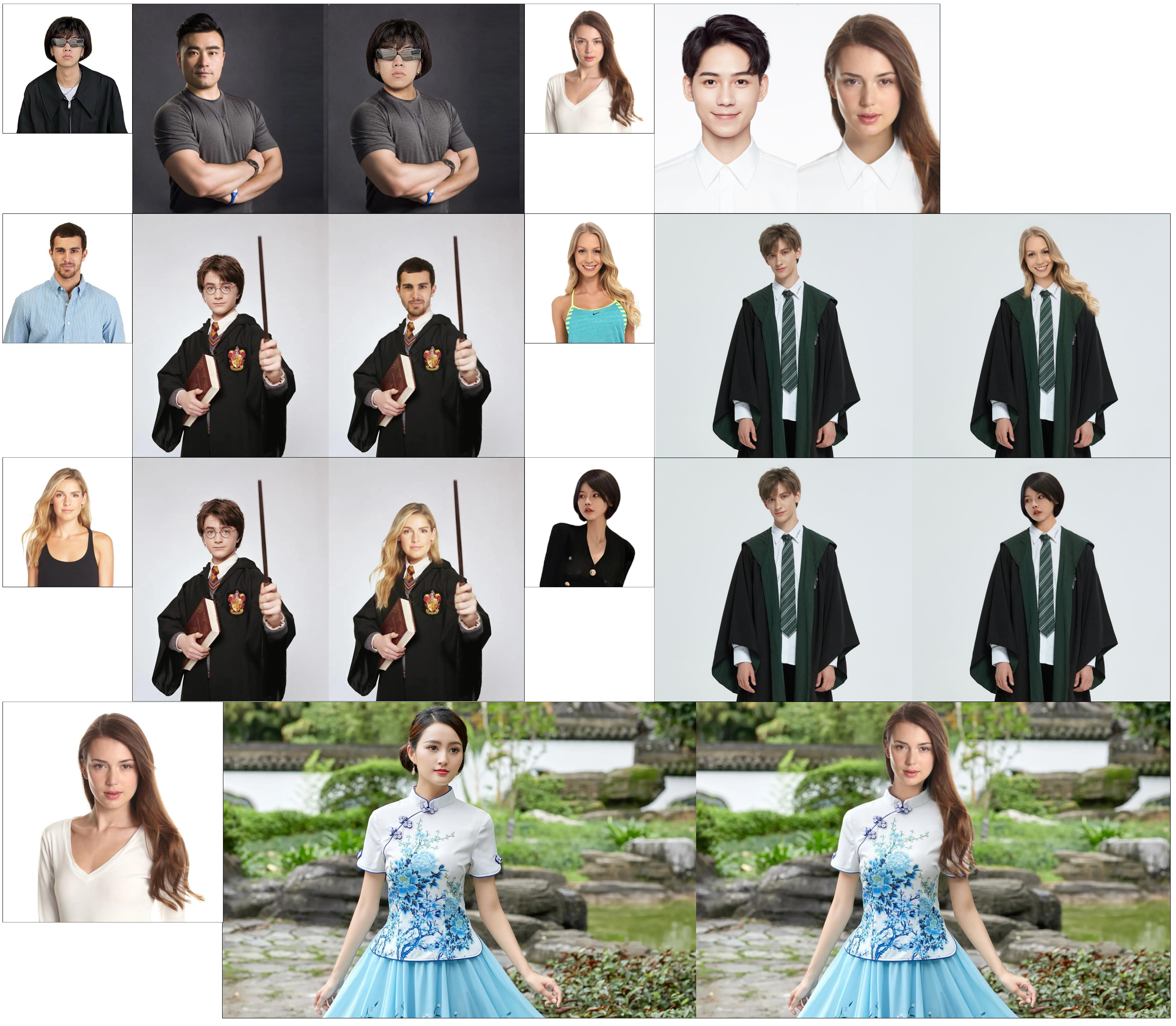}
  \captionof{figure}{Head swapping in the wild. From left to right in each group of images: source head, source body, result.}
  \label{fig:wild}
\end{center}

\section{More results}
\subsection{Head Swapping in the Wild}
To apply our head swapping pipeline in the wild, We employ an image segmentation method PaddleSeg~\cite{liu2021paddleseg} to cut the source head and source body, and achieve a seamless head swapping by our HS-Diffusion. Then we paste the head swapping result onto the background of the source body image, where the missing background region is inpainted by a SOTA inpainting method LaMa~\cite{suvorov2021resolution}. As shown in Fig.~\ref{fig:wild}, the pose and size of human heads are various, so one's neck region is hard to fit other heads. Benefiting from that our semantic-mixing LDM can adaptively generate the transition region (i.e., the neck region and the covered region) to stitch the source head and source body seamlessly, we achieve photo-realistic head swapping in the wild. It has important implications for a variety of applications in commercial and entertainment scenarios, such as occupational
photos composition and cosplay.

\begin{center}
\captionsetup{type=figure}
  \centering
  \includegraphics[width=0.91\linewidth]{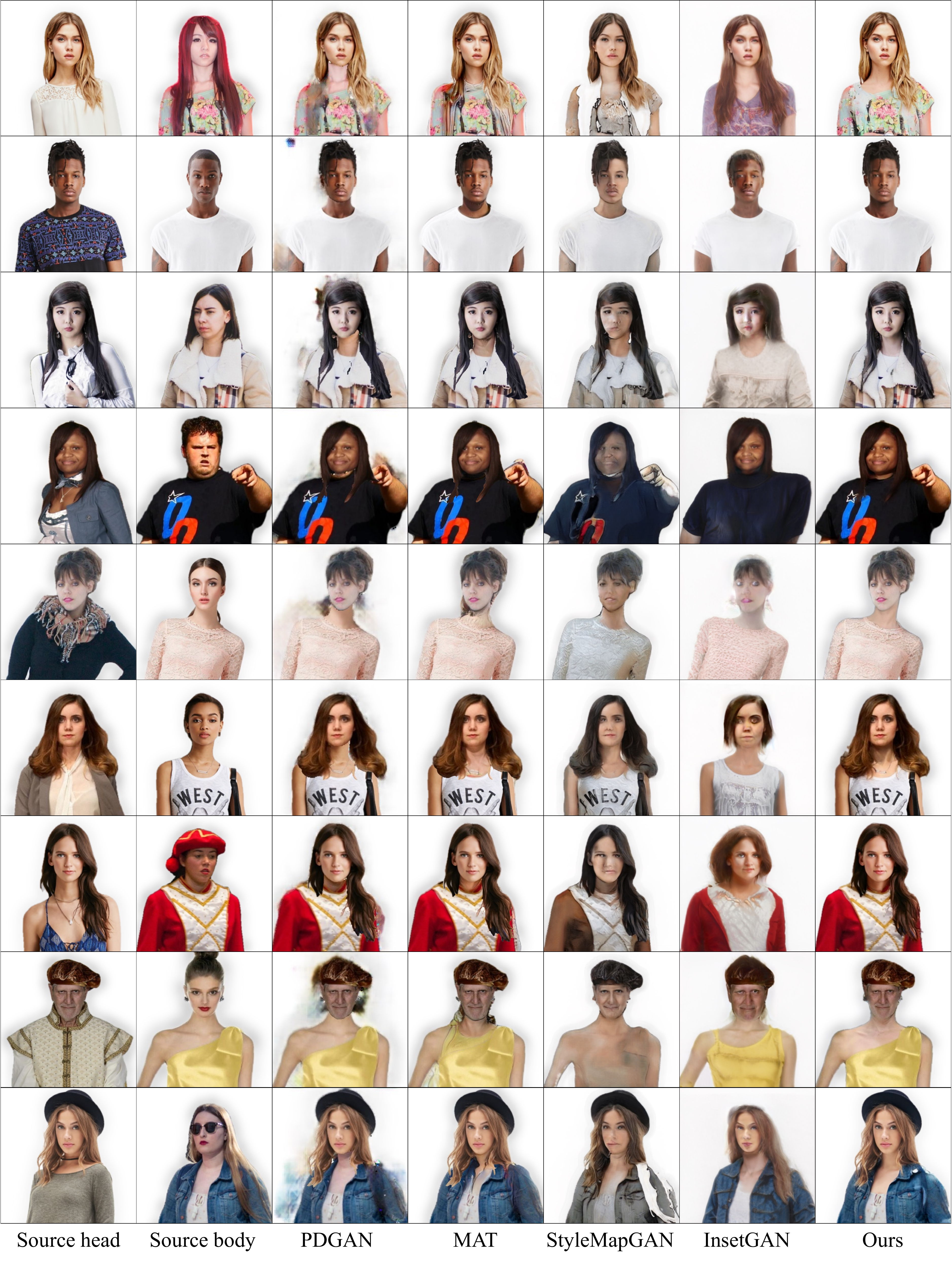}
  \captionof{figure}{More qualitative comparisons with PDGAN~\cite{liu2021pd}, MAT~\cite{li2022mat}, StyleMapGAN~\cite{kim2021exploiting} and InsetGAN~\cite{fruhstuck2022insetgan}.}
  \label{fig:a4}
\end{center}

\begin{algorithm}[H]
\caption{Head swapping pipeline with our semantic-mixing LDM. Given a well-trained LDM~(including a well-trained VQGAN~($\mathcal{E},\mathcal{D}$), a denoising U-Net $\epsilon_\theta$, a condition encoder $\tau_\theta$) and a semantic layout generator~($G_{layout}$).}
\label{al:hs}
\textbf{Input}: Two neck-aligned half-body images~($x_1,x_2$) and their semantic layouts~($l_1,l_2$), the head mask $m^H$, body mask $m^B$ and the rest region mask $m^r=1-m^H-m^B$.\\
\textbf{Output}: A head swapping result $\tilde{x}$.\\
\vspace{-13pt}
\begin{algorithmic}[1] 
\STATE $l_{blend} = l_1\odot m^H + l_2\odot m^B + 0\odot m^r$;\\
\STATE $s = \tau_\theta(G_{layout}(l_{blend}))$;\\
\STATE $z^H_0 = \mathcal{E}(x_1),z^B_0 = \mathcal{E}(x_2)$;\\
\STATE $z_T\sim\mathcal{N}(0,\mathbf{I})$;
\FOR{all $t$ from $T$ to 0}
\STATE $z^H_t\sim\mathcal{N}(\sqrt{\bar{\alpha}_{t}}z^H_0,(1-\bar{\alpha}_t)\mathbf{I})$;\\
\STATE $z^B_t\sim\mathcal{N}(\sqrt{\bar{\alpha}_{t}}z^B_0,(1-\bar{\alpha}_t)\mathbf{I})$;\\
  \STATE $\hat{z_t}=z_t^H\odot m^H + z_t^B\odot m^B + z_t\odot m^r$;\\
  \STATE Denoise to $z_{t-1}$ by Eq.~(2) with $\epsilon_\theta(\hat{z_t}, s)$;\\
\ENDFOR\\
$\tilde{x} = \mathcal{D}(z_0)$;\\
\STATE \textbf{return} $\tilde{x}$
\end{algorithmic}
\end{algorithm}

\begin{table}[h]
\caption{We compute the training time and inference time on the half-body SHHQ256 and half-body SHHQ512 datasets. `d' denotes day and `h' denotes hour.}
\label{tab:time}
\centering
\begin{tabular}{cccccc}
\toprule
Spatial size & 256$^2$       & \multicolumn{2}{c}{256$^2$} & \multicolumn{2}{c}{512$^2$} \\

Model          & $G_{layout}$ & VQGAN      & LDM     & VQGAN      & LDM     \\
\midrule
Batch size     & 24            & 8              & 20         & 1              & 6          \\
Images trained & 1.4M          & 14.7M          & 10.1M      & 7.2M           & 5.5M       \\
Training time  & 2d14h         & 6d3h           & 1d8h       & 16d8h          & 3d11h      \\
Inference time & 0.22s         & 0.08s          & 2.36s      & 0.09s          & 10.07s    \\
\bottomrule
\end{tabular}
\end{table}

\section{Implementation Details}
\subsection{Training Details}

To land the diffusion process into the latent space, we first finetune the autoencoder~\cite{esser2021taming}~(i.e., VQGAN) pretrained on OpenImages~\cite{kuznetsova2020open} with our half-body SHHQ dataset. Secondly, based on the well-trained VQGAN, we train the latent diffusion model~(LDM) with the proposed head-cover augmentation as shown in the Fig.~\ref{fig:train_process}. In addition, we train the $G_{layout}$ with head-cover augmentation and removing the neck region of input layout. We provide the training and inference details in Fig.~\ref{tab:time}, where the LDM and VQGAN are trained on 8 Nvidia V100 GPUs, and $G_{layout}$ is trained on a single Nvidia V100 GPU. Since the semantic layout can be directly resized by nearest neighbor interpolation without side effects, we only train the $G_{layout}$ with our half-body SHHQ256 dataset which can be used for head swapping on half-body SHHQ512. The inference time is averaged by the models' total inference time on the test dataset with 50 DDIM~\cite{song2020denoising} steps.

The introduced human parsing method SCHP~\cite{li2020self} can obtain the semantic layouts of half-body images. The layout includes 20 classes: \textit{background}, \textit{hat}, \textit{hair}, \textit{glove}, \textit{sunglasses}, \textit{upper-clothes}, \textit{dress}, \textit{coat}, \textit{socks}, \textit{pants}, \textit{skin}, \textit{scarf}, \textit{skirt}, \textit{face}, \textit{left-arm}, \textit{right-arm}, \textit{left-leg}, \textit{right-leg}, \textit{left-shoe}, \textit{right-shoe}. We define the \textit{hat}, \textit{hair}, \textit{sunglasses} and \textit{face} as the head region, the \textit{glove}, \textit{upper-clothes}, \textit{dress}, \textit{coat}, \textit{socks}, \textit{pants}, \textit{scarf}, \textit{skirt}, \textit{left-arm}, \textit{right-arm}, \textit{left-leg}, \textit{right-leg}, \textit{left-shoe} and \textit{right-shoe} as the body region.

\subsection{Head Swapping Pipeline}
In addition to the Fig.~\ref{fig:infer_process} and the head swapping steps mentioned in Sec.~\ref{sec:method}, we further describe our head swapping pipeline in Fig.~\ref{al:hs}.

\section{Limitations}
There are two major limitations with our framework: 1) Although we implement the denoising process in latent space, which is faster than the diffusion-based methods in image space, it is still a long way from real-time head swapping. Therefore, we will explore to accelerate this pipeline. 2) The robustness of our framework is improved by training with our head-cover augmentation strategy, but the performance is still affected to some extent by the semantic layout.


\begin{thebibliography}{10}\itemsep=-1pt

\bibitem{albahar2021pose}
Badour AlBahar, Jingwan Lu, Jimei Yang, Zhixin Shu, Eli Shechtman, and Jia-Bin
  Huang.
\newblock Pose with style: Detail-preserving pose-guided image synthesis with
  conditional stylegan.
\newblock {\em ACM Transactions on Graphics (TOG)}, 40(6):1--11, 2021.

\bibitem{avrahami2022blended_1}
Omri Avrahami, Ohad Fried, and Dani Lischinski.
\newblock Blended latent diffusion.
\newblock {\em arXiv preprint arXiv:2206.02779}, 2022.

\bibitem{avrahami2022blended_0}
Omri Avrahami, Dani Lischinski, and Ohad Fried.
\newblock Blended diffusion for text-driven editing of natural images.
\newblock In {\em Proceedings of the IEEE/CVF Conference on Computer Vision and
  Pattern Recognition}, pages 18208--18218, 2022.

\bibitem{brock2018large}
Andrew Brock, Jeff Donahue, and Karen Simonyan.
\newblock Large scale gan training for high fidelity natural image synthesis.
\newblock {\em arXiv preprint arXiv:1809.11096}, 2018.

\bibitem{burt1987laplacian}
Peter~J Burt and Edward~H Adelson.
\newblock The laplacian pyramid as a compact image code.
\newblock In {\em Readings in computer vision}, pages 671--679. Elsevier, 1987.

\bibitem{chan2022efficient}
Eric~R Chan, Connor~Z Lin, Matthew~A Chan, Koki Nagano, Boxiao Pan, Shalini
  De~Mello, Orazio Gallo, Leonidas~J Guibas, Jonathan Tremblay, Sameh Khamis,
  et~al.
\newblock Efficient geometry-aware 3d generative adversarial networks.
\newblock In {\em Proceedings of the IEEE/CVF Conference on Computer Vision and
  Pattern Recognition}, pages 16123--16133, 2022.

\bibitem{chen2020simswap}
Renwang Chen, Xuanhong Chen, Bingbing Ni, and Yanhao Ge.
\newblock Simswap: An efficient framework for high fidelity face swapping.
\newblock In {\em Proceedings of the 28th ACM International Conference on
  Multimedia}, pages 2003--2011, 2020.

\bibitem{deng2019arcface}
Jiankang Deng, Jia Guo, Niannan Xue, and Stefanos Zafeiriou.
\newblock Arcface: Additive angular margin loss for deep face recognition.
\newblock In {\em Proceedings of the IEEE/CVF conference on computer vision and
  pattern recognition}, pages 4690--4699, 2019.

\bibitem{dhariwal2021diffusion}
Prafulla Dhariwal and Alexander Nichol.
\newblock Diffusion models beat gans on image synthesis.
\newblock {\em Advances in Neural Information Processing Systems},
  34:8780--8794, 2021.

\bibitem{esser2021taming}
Patrick Esser, Robin Rombach, and Bjorn Ommer.
\newblock Taming transformers for high-resolution image synthesis.
\newblock In {\em Proceedings of the IEEE/CVF conference on computer vision and
  pattern recognition}, pages 12873--12883, 2021.

\bibitem{fruhstuck2022insetgan}
Anna Fr{\"u}hst{\"u}ck, Krishna~Kumar Singh, Eli Shechtman, Niloy~J Mitra,
  Peter Wonka, and Jingwan Lu.
\newblock Insetgan for full-body image generation.
\newblock In {\em Proceedings of the IEEE/CVF Conference on Computer Vision and
  Pattern Recognition}, pages 7723--7732, 2022.

\bibitem{fu2022stylegan}
Jianglin Fu, Shikai Li, Yuming Jiang, Kwan-Yee Lin, Chen Qian, Chen~Change Loy,
  Wayne Wu, and Ziwei Liu.
\newblock Stylegan-human: A data-centric odyssey of human generation.
\newblock {\em arXiv preprint arXiv:2204.11823}, 2022.

\bibitem{georgopoulos2021mitigating}
Markos Georgopoulos, James Oldfield, Mihalis~A Nicolaou, Yannis Panagakis, and
  Maja Pantic.
\newblock Mitigating demographic bias in facial datasets with style-based
  multi-attribute transfer.
\newblock {\em International Journal of Computer Vision}, 129(7):2288--2307,
  2021.

\bibitem{goel2023pair}
Vidit Goel, Elia Peruzzo, Yifan Jiang, Dejia Xu, Nicu Sebe, Trevor Darrell,
  Zhangyang Wang, and Humphrey Shi.
\newblock Pair-diffusion: Object-level image editing with
  structure-and-appearance paired diffusion models.
\newblock {\em arXiv preprint arXiv:2303.17546}, 2023.

\bibitem{graikos2022diffusion}
Alexandros Graikos, Nikolay Malkin, Nebojsa Jojic, and Dimitris Samaras.
\newblock Diffusion models as plug-and-play priors.
\newblock {\em arXiv preprint arXiv:2206.09012}, 2022.

\bibitem{han2018viton}
Xintong Han, Zuxuan Wu, Zhe Wu, Ruichi Yu, and Larry~S Davis.
\newblock Viton: An image-based virtual try-on network.
\newblock In {\em Proceedings of the IEEE conference on computer vision and
  pattern recognition}, pages 7543--7552, 2018.

\bibitem{ho2020denoising}
Jonathan Ho, Ajay Jain, and Pieter Abbeel.
\newblock Denoising diffusion probabilistic models.
\newblock {\em Advances in Neural Information Processing Systems},
  33:6840--6851, 2020.

\bibitem{ho2020sketch}
Trang-Thi Ho, John~Jethro Virtusio, Yung-Yao Chen, Chih-Ming Hsu, and Kai-Lung
  Hua.
\newblock Sketch-guided deep portrait generation.
\newblock {\em ACM Transactions on Multimedia Computing, Communications, and
  Applications (TOMM)}, 16(3):1--18, 2020.

\bibitem{isogawa2019better}
Mariko Isogawa, Dan Mikami, Kosuke Takahashi, Daisuke Iwai, Kosuke Sato, and
  Hideaki Kimata.
\newblock Which is the better inpainted image? training data generation without
  any manual operations.
\newblock {\em International Journal of Computer Vision}, 127:1751--1766, 2019.

\bibitem{jabbar2021survey}
Abdul Jabbar, Xi Li, and Bourahla Omar.
\newblock A survey on generative adversarial networks: Variants, applications,
  and training.
\newblock {\em ACM Computing Surveys (CSUR)}, 54(8):1--49, 2021.

\bibitem{jang2016categorical}
Eric Jang, Shixiang Gu, and Ben Poole.
\newblock Categorical reparameterization with gumbel-softmax.
\newblock {\em arXiv preprint arXiv:1611.01144}, 2016.

\bibitem{karras2019style}
Tero Karras, Samuli Laine, and Timo Aila.
\newblock A style-based generator architecture for generative adversarial
  networks.
\newblock In {\em Proceedings of the IEEE/CVF conference on computer vision and
  pattern recognition}, pages 4401--4410, 2019.

\bibitem{karras2020analyzing}
Tero Karras, Samuli Laine, Miika Aittala, Janne Hellsten, Jaakko Lehtinen, and
  Timo Aila.
\newblock Analyzing and improving the image quality of stylegan.
\newblock In {\em Proceedings of the IEEE/CVF conference on computer vision and
  pattern recognition}, pages 8110--8119, 2020.

\bibitem{kazemi2014one}
Vahid Kazemi and Josephine Sullivan.
\newblock One millisecond face alignment with an ensemble of regression trees.
\newblock In {\em Proceedings of the IEEE conference on computer vision and
  pattern recognition}, pages 1867--1874, 2014.

\bibitem{kim2021exploiting}
Hyunsu Kim, Yunjey Choi, Junho Kim, Sungjoo Yoo, and Youngjung Uh.
\newblock Exploiting spatial dimensions of latent in gan for real-time image
  editing.
\newblock In {\em Proceedings of the IEEE/CVF Conference on Computer Vision and
  Pattern Recognition}, pages 852--861, 2021.

\bibitem{kim2022smooth}
Jiseob Kim, Jihoon Lee, and Byoung-Tak Zhang.
\newblock Smooth-swap: A simple enhancement for face-swapping with smoothness.
\newblock In {\em Proceedings of the IEEE/CVF Conference on Computer Vision and
  Pattern Recognition}, pages 10779--10788, 2022.

\bibitem{kingma2014adam}
Diederik~P Kingma and Jimmy Ba.
\newblock Adam: A method for stochastic optimization.
\newblock {\em arXiv preprint arXiv:1412.6980}, 2014.

\bibitem{krizhevsky2012imagenet}
Alex Krizhevsky, Ilya Sutskever, and Geoffrey~E Hinton.
\newblock Imagenet classification with deep convolutional neural networks.
\newblock {\em Advances in neural information processing systems},
  25:1097--1105, 2012.

\bibitem{kuznetsova2020open}
Alina Kuznetsova, Hassan Rom, Neil Alldrin, Jasper Uijlings, Ivan Krasin, Jordi
  Pont-Tuset, Shahab Kamali, Stefan Popov, Matteo Malloci, Alexander
  Kolesnikov, et~al.
\newblock The open images dataset v4.
\newblock {\em International Journal of Computer Vision}, 128(7):1956--1981,
  2020.

\bibitem{lee2020maskgan}
Cheng-Han Lee, Ziwei Liu, Lingyun Wu, and Ping Luo.
\newblock Maskgan: Towards diverse and interactive facial image manipulation.
\newblock In {\em Proceedings of the IEEE/CVF Conference on Computer Vision and
  Pattern Recognition}, pages 5549--5558, 2020.

\bibitem{lee2020drit++}
Hsin-Ying Lee, Hung-Yu Tseng, Qi Mao, Jia-Bin Huang, Yu-Ding Lu, Maneesh Singh,
  and Ming-Hsuan Yang.
\newblock Drit++: Diverse image-to-image translation via disentangled
  representations.
\newblock {\em International Journal of Computer Vision}, 128:2402--2417, 2020.

\bibitem{li2021toward}
Kedan Li, Min~Jin Chong, Jeffrey Zhang, and Jingen Liu.
\newblock Toward accurate and realistic outfits visualization with attention to
  details.
\newblock In {\em Proceedings of the IEEE/CVF Conference on Computer Vision and
  Pattern Recognition}, pages 15546--15555, 2021.

\bibitem{li2019faceshifter}
Lingzhi Li, Jianmin Bao, Hao Yang, Dong Chen, and Fang Wen.
\newblock Faceshifter: Towards high fidelity and occlusion aware face swapping.
\newblock {\em arXiv preprint arXiv:1912.13457}, 2019.

\bibitem{li2020self}
Peike Li, Yunqiu Xu, Yunchao Wei, and Yi Yang.
\newblock Self-correction for human parsing.
\newblock {\em IEEE Transactions on Pattern Analysis and Machine Intelligence},
  2020.

\bibitem{li2022mat}
Wenbo Li, Zhe Lin, Kun Zhou, Lu Qi, Yi Wang, and Jiaya Jia.
\newblock Mat: Mask-aware transformer for large hole image inpainting.
\newblock In {\em Proceedings of the IEEE/CVF Conference on Computer Vision and
  Pattern Recognition}, pages 10758--10768, 2022.

\bibitem{Li_Huang_Cao_He_Tan_2020}
Yi Li, Huaibo Huang, Jie Cao, Ran He, and Tieniu Tan.
\newblock Disentangled representation learning of makeup portraits in the wild.
\newblock {\em International Journal of Computer Vision}, page 2166–2184, Sep
  2020.

\bibitem{li2023gligen}
Yuheng Li, Haotian Liu, Qingyang Wu, Fangzhou Mu, Jianwei Yang, Jianfeng Gao,
  Chunyuan Li, and Yong~Jae Lee.
\newblock Gligen: Open-set grounded text-to-image generation.
\newblock In {\em Proceedings of the IEEE/CVF Conference on Computer Vision and
  Pattern Recognition}, pages 22511--22521, 2023.

\bibitem{liu2021pd}
Hongyu Liu, Ziyu Wan, Wei Huang, Yibing Song, Xintong Han, and Jing Liao.
\newblock Pd-gan: Probabilistic diverse gan for image inpainting.
\newblock In {\em Proceedings of the IEEE/CVF Conference on Computer Vision and
  Pattern Recognition}, pages 9371--9381, 2021.

\bibitem{liu2022semantic}
Xian Liu, Yinghao Xu, Qianyi Wu, Hang Zhou, Wayne Wu, and Bolei Zhou.
\newblock Semantic-aware implicit neural audio-driven video portrait
  generation.
\newblock {\em arXiv preprint arXiv:2201.07786}, 2022.

\bibitem{liu2021paddleseg}
Yi Liu, Lutao Chu, Guowei Chen, Zewu Wu, Zeyu Chen, Baohua Lai, and Yuying Hao.
\newblock Paddleseg: A high-efficient development toolkit for image
  segmentation, 2021.

\bibitem{lu2022dpm}
Cheng Lu, Yuhao Zhou, Fan Bao, Jianfei Chen, Chongxuan Li, and Jun Zhu.
\newblock Dpm-solver: A fast ode solver for diffusion probabilistic model
  sampling in around 10 steps.
\newblock {\em arXiv preprint arXiv:2206.00927}, 2022.

\bibitem{lucic2017gans}
Mario Lucic, Karol Kurach, Marcin Michalski, Sylvain Gelly, and Olivier
  Bousquet.
\newblock Are gans created equal? a large-scale study.
\newblock {\em arXiv preprint arXiv:1711.10337}, 2017.

\bibitem{mao2017least}
Xudong Mao, Qing Li, Haoran Xie, Raymond~YK Lau, Zhen Wang, and Stephen
  Paul~Smolley.
\newblock Least squares generative adversarial networks.
\newblock In {\em Proceedings of the IEEE international conference on computer
  vision}, pages 2794--2802, 2017.

\bibitem{Moschoglou_Ploumpis_Nicolaou_Papaioannou_Zafeiriou_2020}
Stylianos Moschoglou, Stylianos Ploumpis, Mihalis~A. Nicolaou, Athanasios
  Papaioannou, and Stefanos Zafeiriou.
\newblock 3dfacegan: Adversarial nets for 3d face representation, generation,
  and translation.
\newblock {\em International Journal of Computer Vision}, page 2534–2551, Nov
  2020.

\bibitem{mou2023t2i}
Chong Mou, Xintao Wang, Liangbin Xie, Jian Zhang, Zhongang Qi, Ying Shan, and
  Xiaohu Qie.
\newblock T2i-adapter: Learning adapters to dig out more controllable ability
  for text-to-image diffusion models.
\newblock {\em arXiv preprint arXiv:2302.08453}, 2023.

\bibitem{nichol2021improved}
Alexander~Quinn Nichol and Prafulla Dhariwal.
\newblock Improved denoising diffusion probabilistic models.
\newblock In {\em International Conference on Machine Learning}, pages
  8162--8171. PMLR, 2021.

\bibitem{nizan2020breaking}
Ori Nizan and Ayellet Tal.
\newblock Breaking the cycle-colleagues are all you need.
\newblock In {\em Proceedings of the IEEE/CVF Conference on Computer Vision and
  Pattern Recognition}, pages 7860--7869, 2020.

\bibitem{perov2020deepfacelab}
Ivan Perov, Daiheng Gao, Nikolay Chervoniy, Kunlin Liu, Sugasa Marangonda,
  Chris Um{\'e}, Mr Dpfks, Carl~Shift Facenheim, Luis RP, Jian Jiang, et~al.
\newblock Deepfacelab: Integrated, flexible and extensible face-swapping
  framework.
\newblock {\em arXiv preprint arXiv:2005.05535}, 2020.

\bibitem{qin2020u2}
Xuebin Qin, Zichen Zhang, Chenyang Huang, Masood Dehghan, Osmar~R Zaiane, and
  Martin Jagersand.
\newblock U2-net: Going deeper with nested u-structure for salient object
  detection.
\newblock {\em Pattern recognition}, 106:107404, 2020.

\bibitem{rombach2022high}
Robin Rombach, Andreas Blattmann, Dominik Lorenz, Patrick Esser, and Bj{\"o}rn
  Ommer.
\newblock High-resolution image synthesis with latent diffusion models.
\newblock In {\em Proceedings of the IEEE/CVF Conference on Computer Vision and
  Pattern Recognition}, pages 10684--10695, 2022.

\bibitem{saharia2022photorealistic}
Chitwan Saharia, William Chan, Saurabh Saxena, Lala Li, Jay Whang, Emily~L
  Denton, Kamyar Ghasemipour, Raphael Gontijo~Lopes, Burcu Karagol~Ayan, Tim
  Salimans, et~al.
\newblock Photorealistic text-to-image diffusion models with deep language
  understanding.
\newblock {\em Advances in Neural Information Processing Systems},
  35:36479--36494, 2022.

\bibitem{shu2022few}
Changyong Shu, Hemao Wu, Hang Zhou, Jiaming Liu, Zhibin Hong, Changxing Ding,
  Junyu Han, Jingtuo Liu, Errui Ding, and Jingdong Wang.
\newblock Few-shot head swapping in the wild.
\newblock In {\em Proceedings of the IEEE/CVF Conference on Computer Vision and
  Pattern Recognition}, pages 10789--10798, 2022.

\bibitem{song2020denoising}
Jiaming Song, Chenlin Meng, and Stefano Ermon.
\newblock Denoising diffusion implicit models.
\newblock {\em arXiv preprint arXiv:2010.02502}, 2020.

\bibitem{su2022drawinginstyles}
Wanchao Su, Hui Ye, Shu-Yu Chen, Lin Gao, and Hongbo Fu.
\newblock Drawinginstyles: Portrait image generation and editing with spatially
  conditioned stylegan.
\newblock {\em IEEE Transactions on Visualization and Computer Graphics}, 2022.

\bibitem{suvorov2021resolution}
Roman Suvorov, Elizaveta Logacheva, Anton Mashikhin, Anastasia Remizova,
  Arsenii Ashukha, Aleksei Silvestrov, Naejin Kong, Harshith Goka, Kiwoong
  Park, and Victor Lempitsky.
\newblock Resolution-robust large mask inpainting with fourier convolutions.
\newblock {\em arXiv preprint arXiv:2109.07161}, 2021.

\bibitem{tov2021designing}
Omer Tov, Yuval Alaluf, Yotam Nitzan, Or Patashnik, and Daniel Cohen-Or.
\newblock Designing an encoder for stylegan image manipulation.
\newblock {\em ACM Transactions on Graphics (TOG)}, 40(4):1--14, 2021.

\bibitem{wang2022semantic}
Weilun Wang, Jianmin Bao, Wengang Zhou, Dongdong Chen, Dong Chen, Lu Yuan, and
  Houqiang Li.
\newblock Semantic image synthesis via diffusion models.
\newblock {\em arXiv preprint arXiv:2207.00050}, 2022.

\bibitem{wang2022latent}
Yaohui Wang, Di Yang, Francois Bremond, and Antitza Dantcheva.
\newblock Latent image animator: Learning to animate images via latent space
  navigation.
\newblock {\em arXiv preprint arXiv:2203.09043}, 2022.

\bibitem{wang2004image}
Zhou Wang, Alan~C Bovik, Hamid~R Sheikh, and Eero~P Simoncelli.
\newblock Image quality assessment: from error visibility to structural
  similarity.
\newblock {\em IEEE transactions on image processing}, 13(4):600--612, 2004.

\bibitem{wu2019deep}
Xian Wu, Rui-Long Li, Fang-Lue Zhang, Jian-Cheng Liu, Jue Wang, Ariel Shamir,
  and Shi-Min Hu.
\newblock Deep portrait image completion and extrapolation.
\newblock {\em IEEE Transactions on Image Processing}, 29:2344--2355, 2019.

\bibitem{xia2022gan}
Weihao Xia, Yulun Zhang, Yujiu Yang, Jing-Hao Xue, Bolei Zhou, and Ming-Hsuan
  Yang.
\newblock Gan inversion: A survey.
\newblock {\em IEEE Transactions on Pattern Analysis and Machine Intelligence},
  45(3):3121--3138, 2022.

\bibitem{xu2022region}
Chao Xu, Jiangning Zhang, Miao Hua, Qian He, Zili Yi, and Yong Liu.
\newblock Region-aware face swapping.
\newblock In {\em Proceedings of the IEEE/CVF Conference on Computer Vision and
  Pattern Recognition}, pages 7632--7641, 2022.

\bibitem{xu2022high}
Yangyang Xu, Bailin Deng, Junle Wang, Yanqing Jing, Jia Pan, and Shengfeng He.
\newblock High-resolution face swapping via latent semantics disentanglement.
\newblock In {\em Proceedings of the IEEE/CVF Conference on Computer Vision and
  Pattern Recognition}, pages 7642--7651, 2022.

\bibitem{xu2021facecontroller}
Zhiliang Xu, Xiyu Yu, Zhibin Hong, Zhen Zhu, Junyu Han, Jingtuo Liu, Errui
  Ding, and Xiang Bai.
\newblock Facecontroller: Controllable attribute editing for face in the wild.
\newblock In {\em AAAI}, 2021.

\bibitem{yang2020towards}
Han Yang, Ruimao Zhang, Xiaobao Guo, Wei Liu, Wangmeng Zuo, and Ping Luo.
\newblock Towards photo-realistic virtual try-on by adaptively
  generating-preserving image content.
\newblock In {\em Proceedings of the IEEE/CVF conference on computer vision and
  pattern recognition}, pages 7850--7859, 2020.

\bibitem{zhang2023adding}
Lvmin Zhang and Maneesh Agrawala.
\newblock Adding conditional control to text-to-image diffusion models.
\newblock {\em arXiv preprint arXiv:2302.05543}, 2023.

\bibitem{zhang2018perceptual}
Richard Zhang, Phillip Isola, Alexei~A Efros, Eli Shechtman, and Oliver Wang.
\newblock The unreasonable effectiveness of deep features as a perceptual
  metric.
\newblock In {\em CVPR}, 2018.

\bibitem{zhang2023sadtalker}
Wenxuan Zhang, Xiaodong Cun, Xuan Wang, Yong Zhang, Xi Shen, Yu Guo, Ying Shan,
  and Fei Wang.
\newblock Sadtalker: Learning realistic 3d motion coefficients for stylized
  audio-driven single image talking face animation.
\newblock In {\em Proceedings of the IEEE/CVF Conference on Computer Vision and
  Pattern Recognition}, pages 8652--8661, 2023.

\bibitem{zhang2023multi}
Xuanmeng Zhang, Zhedong Zheng, Daiheng Gao, Bang Zhang, Yi Yang, and Tat-Seng
  Chua.
\newblock Multi-view consistent generative adversarial networks for
  compositional 3d-aware image synthesis.
\newblock {\em International Journal of Computer Vision}, pages 1--24, 2023.

\bibitem{zhao2020layout2image}
Bo Zhao, Weidong Yin, Lili Meng, and Leonid Sigal.
\newblock Layout2image: Image generation from layout.
\newblock {\em International Journal of Computer Vision}, 128:2418--2435, 2020.

\bibitem{zhu2021one}
Yuhao Zhu, Qi Li, Jian Wang, Cheng-Zhong Xu, and Zhenan Sun.
\newblock One shot face swapping on megapixels.
\newblock In {\em Proceedings of the IEEE/CVF conference on computer vision and
  pattern recognition}, pages 4834--4844, 2021.

\end{thebibliography}
\end{document}